\documentclass[10pt,twocolumn,letterpaper]{article}

\usepackage{iccv}
\usepackage{times}
\usepackage{epsfig}
\usepackage{graphicx}
\usepackage{amsmath}
\usepackage{amssymb}
\usepackage{booktabs}
\usepackage{multirow}
\usepackage[table]{xcolor}
\usepackage{color}
\usepackage{subfig}
\usepackage[misc]{ifsym} 

\definecolor{linkcolor}{RGB}{255,0,0}
\definecolor{urlcolor}{RGB}{255,105,180}
\definecolor{citecolor}{RGB}{66,168,235}

\definecolor{linkcolor}{RGB}{255,0,0}
\definecolor{urlcolor}{RGB}{255,105,180}
\definecolor{citecolor}{RGB}{66,168,235}
\usepackage[pagebackref=true,breaklinks=true,colorlinks=true,bookmarks=false,linkcolor=linkcolor,urlcolor=urlcolor,citecolor=citecolor]{hyperref}

\iccvfinalcopy 

\def\iccvPaperID{5113} 

\ificcvfinal\fi

\begin{document}

\title{Towards Robust Referring Image Segmentation}

\author{
Jianzong Wu$^{1}$ \quad
Xiangtai Li$^{1 \textrm{\Letter}}$  \quad
Xia Li $^{2}$ \quad
Henghui Ding$^{2}$ \quad
Yunhai Tong$^{1}$ \quad
Dacheng Tao$^{3}$
\\[0.1cm]
\small $ ^1$ School of Artificial Intelligence, Key Laboratory of Machine Perception (MOE), Peking University \\
\small $ ^2$ ETH Zurich \quad \small $ ^3$ The University of Sydney \\
\small jzwu@stu.pku.edu.cn \quad  lxtpku@pku.edu.cn \\
}

\ificcvfinal\thispagestyle{empty}\fi

\twocolumn[{%
\renewcommand\twocolumn[1][]{#1}%
\maketitle 
\centering 
\includegraphics[width=0.95\textwidth]{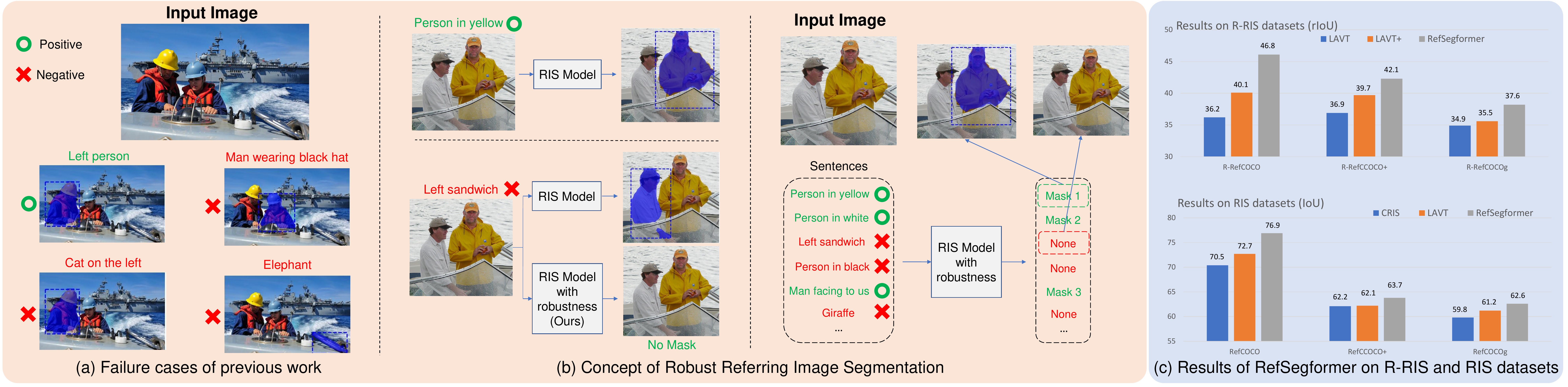} 
\captionof{figure}{Illustration of R-RIS. (a) Failure cases of a RIS model \cite{yu2018mattnet} given negative sentences. (b) Concept of the proposed R-RIS task. A robust RIS model should output \textbf{no mask} for negative inputs and correct masks for positive inputs.(c) Our proposed ReferSegformer achieves stronger performance over six datasets for RIS and R-RIS.} 
\label{fig:teaser}
\vspace{3mm}
}]

\begin{abstract}
Referring Image Segmentation (RIS) is a fundamental vision-language task that outputs object masks based on text descriptions. 
Many works have achieved considerable progress for RIS, including different fusion method designs.
In this work, we explore an essential question, ``What if the text description is wrong or misleading?'' For example, the described objects are not in the image. We term such a sentence as a negative sentence. 
However, existing solutions for RIS cannot handle such a setting. 
To this end, we propose a new formulation of RIS, named Robust Referring Image Segmentation (R-RIS). 
It considers the negative sentence inputs besides the regular positive text inputs. 
To facilitate this new task, we create three R-RIS datasets by augmenting existing RIS datasets with negative sentences and propose new metrics to evaluate both types of inputs in a unified manner. 
Furthermore, we propose a new transformer-based model, called RefSegformer, with a token-based vision and language fusion module. Our design can be easily extended to our R-RIS setting by adding extra blank tokens. 
Our proposed RefSegformer achieves state-of-the-art results on both RIS and R-RIS datasets, establishing a solid baseline for both settings. 
Our project page is at \url{https://github.com/jianzongwu/robust-ref-seg}.

\end{abstract}

\vspace{-4mm}
\section{Introduction}

Referring Image Segmentation (RIS) requires the model to output object masks in an image based on a given text expression. 
Many previous works~\cite{CGAN, CMPC, ding2022vlt, EFN, VLT, LAVT} have proposed various methods and achieved significant improvements in this field. 
There are also related works on video-level referring segmentation \cite{khoreva2018video, botach2022end, wu2022language, seo2020urvos, liang2021clawcranenet}, which extend RIS to the temporal domain.

However, the RIS task assumes that the referred object always exists in the image. 
Under that concern, a RIS model only needs to locate and segment the objects without verifying whether the text description matches the image content. 
This may limit RIS models' \textit{robustness} and \textit{interpretability}.
Moreover, several works~\cite{geirhos2018generalisation,neyshabur2017exploring,rice2020overfitting,kamann2020benchmarking,zhou2022understanding} explore the robustness and interpretability of vision models for image classification and segmentation.
%
Motivated by that, we argue that a better RIS task should be able to handle both matching and mismatching image-text pairs.

In this work, we rethink the RIS setting by asking an essential question: ``what are the segmentation results given a text description that refers to \textbf{a non-existing object or even a wrong object?}". 
We argue that a robust RIS model should \textbf{not} output any masks for such text descriptions. 
However, existing two-stage methods, \eg, MATTNet~\cite{yu2018mattnet}, acquire segmentation masks from an off-the-shelf image segmentation model and use the text description to select the best matching one, thus can not produce the correct result (no mask) for a reference that describes a non-existing object.
One-stage methods~\cite{VLT, LAVT} also suffer from data bias (referred objects always exist). They tend to output masks even if no such object exists in the image. 
As shown in Fig.~\ref{fig:teaser} (a), there are several failure cases for existing methods. 
A person is segmented as a cat, while a non-existing black hat man is also segmented as a false positive example. 

To address these issues, we \textit{redefine} the RIS task with a concept: \textbf{R}obust \textbf{R}eferring \textbf{I}mage \textbf{S}egmentation (R-RIS). It considers the cases where the given expression may refer to objects not in the images. As shown in Fig~\ref{fig:teaser} (b), compared with the previous RIS, R-RIS requires the model to output blank masks for non-existing objects and regular masks for existing objects. Thus, the R-RIS covers both regular and noisy text inputs. To facilitate the R-RIS research, we build new R-RIS benchmarks based on the existing RIS datasets, including RefCOCO~\cite{RefCOCO}, RefCOCO+~\cite{RefCOCO}, and RefCOCOg~\cite{RefCOCOg, RefCOCOg2}.
In particular, we present five ways to generate false text descriptions, including randomly replacing sentences, replacing names, changing the target objects, changing the attributes, and changing the relationships. 
Moreover, we present two metrics: robust IoU (rIoU) and mean Robust Recall (mRR). 
The former measures the robustness and accuracy of R-RIS models at the pixel level, while the latter measures the instance-level robustness for negative text inputs.

To address the R-RIS task, a naive solution is to use a two-stage approach that combines a standard RIS model with an additional binary classifier to determine the existence of the referred object in the image. The model produces a mask only if the classifier predicts a positive result. 
However, this simple baseline, which relies on existing RIS models not tailored for negative inputs, performs poorly on the R-RIS task. 
To overcome this limitation, we propose RefSegformer, a transformer-based model designed explicitly for the R-RIS task and can serve as a strong baseline for future work.

We introduce RefSegformer, a Transformer-based model that consists of a language encoder, a vision encoder, and an encoder-fusion meta-architecture. The key component of our model is the Vision-Language Token Fusion (VLTF) module, which dynamically selects the most relevant language information using memory tokens and fuses it with the vision encoder using Multi-Head Cross Attention (MHCA). To handle the R-RIS task, we propose a blank token design to indicate the presence or absence of the referred object in the image. The blank tokens are not involved in the MHCA fusion with language inputs, thus avoiding the over-fitting to either positive or negative inputs. RefSegformer achieves state-of-the-art results on \textbf{both} RIS and R-RIS datasets. We conduct extensive experiments and analyses to demonstrate the effectiveness of RefSegformer.

Our main contributions are summarized as follows:
\begin{itemize}
    \item We introduce the R-RIS task, the first robust-based RIS, which extends the RIS task to account for the possibility of incorrect referring expressions.
    \item We construct three benchmarks for the R-RIS task and propose new metrics to evaluate R-RIS models. We also benchmark five baseline methods for R-RIS.
    \item We develop RefSegformer, a Transformer-based model incorporating VLTF modules specially designed for the R-RIS task.
    \item RefSegformer achieves state-of-the-art performance on six datasets from both RIS and R-RIS tasks, demonstrating its effectiveness and robustness.
\end{itemize}
\section{Related Work}


\noindent
\textbf{Referring Image Segmentation.}
This task aims to segment an object that matches a given reference expression. There are mainly two approaches: \textit{decoder-fusion} and \textit{encoder-fusion}. The former~\cite{yu2018mattnet, CMPC, MCN, BUSNet, CGAN, VLT, CRIS, hui2020linguistic, jain2021comprehensive} extracts vision and language features separately and fuses them in a multi-modal decoder. For instance, MAttNet~\cite{yu2018mattnet} uses a two-stage framework that first generates candidate proposals using a Mask R-CNN model~\cite{Mask-RCNN}, then selects the most relevant one based on the linguistic feature of the expression. CMPC~\cite{CMPC} adopts a progressive strategy based on informative words in the expression. The encoder fusion approaches~\cite{EFN, LAVT, li2021mail,liang2022local} fuse language features into vision features early in the vision encoder, which is more prevalent in recent works. Specifically, EFN~\cite{EFN} transforms a CNN backbone into a multi-modal network that uses language to refine the multi-modal features. LAVT~\cite{LAVT}, inspired by the success of vision Transformers, designs pixel-word attention and a language pathway module to fuse linguistic features into a vision Transformer backbone and adopt various feature fusion methods~\cite{swin,VIT,xie2021segformer,Li2022SFNetFA,li2020gated,li2020semantic} in the encoder stage. To summarize, the vision and language encoders can have different designs. However, a common assumption is that the text description refers to an object in the image. In this paper, we challenge this assumption and introduce a new R-RIS task, which requires models to handle both positive and negative information effectively in a more realistic and difficult setting.

\noindent
\textbf{Robustness In Segmentation and Classification.}  Robustness studies on CNNs have been conducted in various benchmarks~\cite{geirhos2018generalisation,neyshabur2017exploring}. Recent works also investigated how to evaluate and improve the robustness of CNNs against different weather conditions~\cite{chen2018domain,sakaridis2019guided,sakaridis2018semantic}, and proposed other image corruption models for benchmarking~\cite{vasiljevic2016examining,hendrycks2019benchmarking,kamann2020benchmarking,altindis2021benchmarking,yamada2022does,zhou2022understanding}.
However, those works mainly focus on the robustness analysis of image-only models for classification and segmentation tasks.
In this paper, we introduce a new cross-modality setting to evaluate the robustness of referring segmentation models where the input texts may not match the image content, which is complementary to previous works.

\noindent
\textbf{Vision Transformer.} Transformer has been widely adopted in computer vision for two primary purposes: the backbone for feature extraction~\cite{VIT,zhang2021analogous,deit_vit,swin,zhang2022eatformer,zhang2023rethinking,li2023transformer,wu2023betrayed,li2022videoknet,li2022panopticpartformer,li2023panopticpartformer++,xu2022fashionformer,cheng2021maskformer} and query modeling for specific tasks~\cite{LAVT,detr}. The former demonstrates the benefits of capturing long-range dependencies among image patch features, while the latter formulates instance-level prediction as a set prediction problem. Furthermore, several works~\cite{gabeur2020mmt, kamath2021mdetr, CLIP} exploit the power of multi-modal modeling with Transformer. Our work also leverages multi-modal vision transformer architecture to develop an effective R-RIS baseline model.


\begin{figure}[!t]
	\centering
	\includegraphics[width=1.0\linewidth]{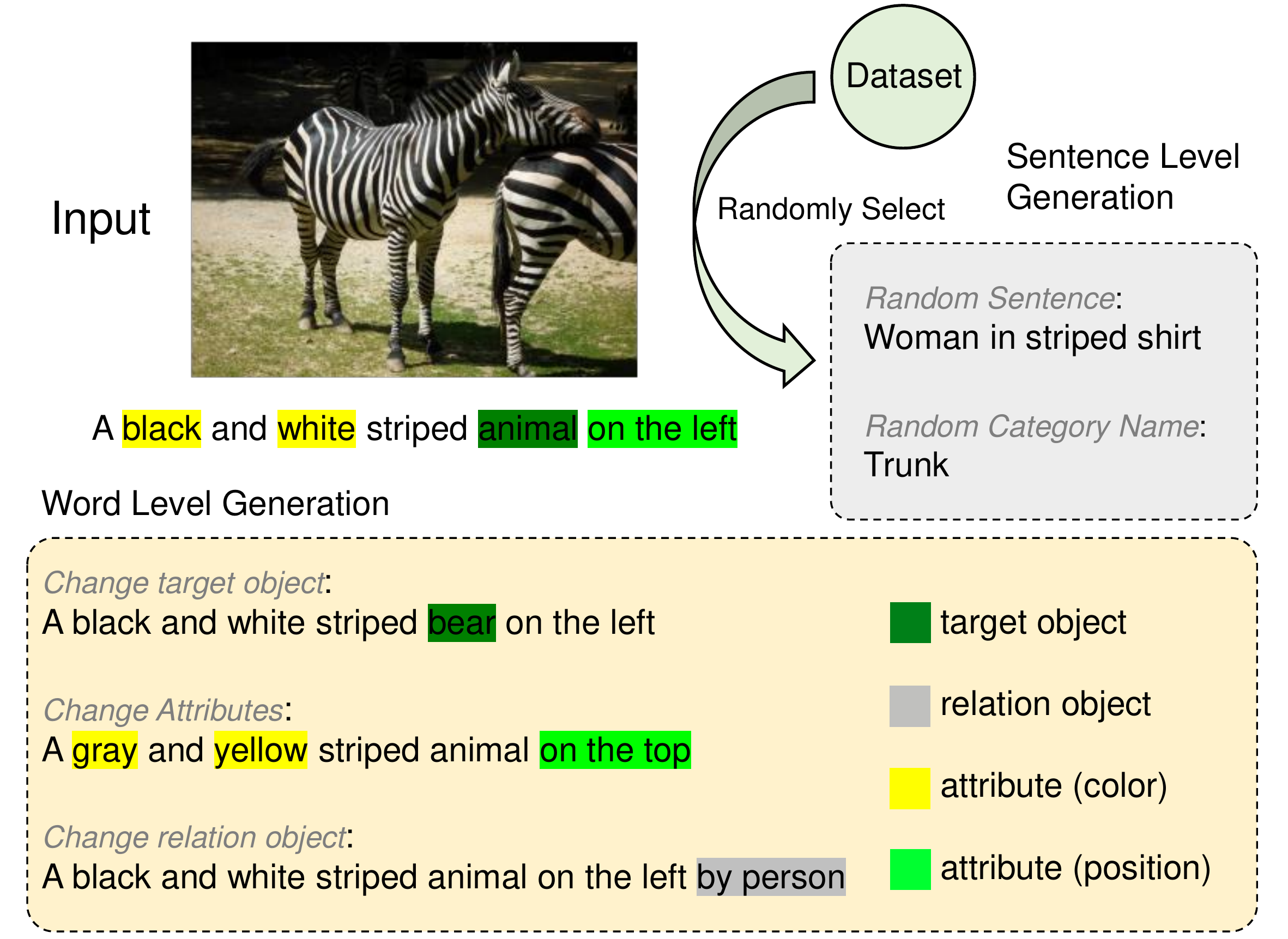}
	\caption{ Illustration of the negative sentence generating process. We propose five different strategies to generate negative sentences from existing descriptions.}
	\label{fig:benchmark}
\end{figure}

\begin{table}[!t]
   \centering
   \scalebox{0.90}{
   \setlength{\tabcolsep}{1.mm}{\begin{tabular}{lccc}
      \toprule[1pt]
      \multirow{2}{*}{Dataset} & Reference & Sentence per & Negative Sentence \\
      & Number & Reference & per Reference \\
      \hline
      \rowcolor{gray!5} RefCOCO & 42,404/3,811 & 2.84/2.84 & -/- \\
      \rowcolor{blue!5} \textbf{R-RefCOCO} & \textbf{42,404/3,811} & \textbf{5.69/13.70} & \textbf{2.84/10.87} \\
      \rowcolor{gray!5} RefCOCO+ & 42,278/3,805 & 2.84/2.83 & -/- \\
      \rowcolor{blue!5} \textbf{R-RefCOCO+} & \textbf{42,278/3,805} & \textbf{5.68/13.04} & \textbf{2.84/10.21} \\
      \rowcolor{gray!5} RefCOCOg & 42,226/2,573 & 1.90/1.90 & -/- \\
      \rowcolor{blue!5} \textbf{R-RefCOCOg} & \textbf{42,226/2,573} & \textbf{3.81/13.20} & \textbf{1.90/11.30} \\
      \bottomrule[1pt]
   \end{tabular}}}
   \caption{ Data statistics of R-RIS datasets and comparison with corresponding RIS datasets. The left numbers refer to the train set, and the right numbers refer to the validation set.}
   \label{tab:data_statistics}
\end{table}

\section{Robust Referring Segmentation Benchmark}

\noindent
\textbf{Overview.} We first introduce the goal of our benchmark. Then we present the detailed process of generating negative sentences. Finally, we introduce new metrics for R-RIS evaluation.  

\noindent
\textbf{Motivation and Concepts.} To evaluate our R-RIS model, we create datasets that contain negative referring expressions, which are text descriptions of objects that do not exist or are incorrectly described in the image. The default RIS expressions are \textit{positive sentences}. Given an input image with a negative sentence as the referring expression, an R-RIS model is expected to output an empty mask, which indicates that no object matches the description. 

\noindent
\textbf{Generating Negative Sentences.} We propose five different methods to generate negative sentences. We describe these methods in detail below:

(1) Randomly select a sentence from another reference in the RIS dataset that does not describe any object in the image. Since all the images come from COCO dataset~\cite{COCO_dataset}, we use COCO annotations to filter out sentences that match any of the object categories in the image. However, some sentences are too vague to be filtered out, such as \textit{``Left one''}, \textit{``Second from left''}, or \textit{``The black''}. These sentences may refer to different objects depending on the image context. To avoid these false negative sentences, we create a list of vague words and exclude sentences that only contain words from this list.

(2) Randomly choose a category name in COCO as a negative sentence. We ensure the image does not contain any object of this category by using the COCO annotations.

(3) Replace the target object in the positive sentence with a randomly chosen category name. For example, \textit{``Man in the left''} becomes \textit{``Cat in the left''} while there is no cat in the image. We use the Natural Language Toolkit (NLTK) to identify nouns in positive sentences and take the first noun as the target object. 

(4) Change adjective words of the target object. For example, \textit{``Man in blue hat''} becomes \textit{``Man in black hat''}. We mainly focus on two types of adjective words: positions and colors, the most common features mentioned in the referring expressions. If the original positive sentence does not have adjective words, we add a random position and color for them. 

(5) Change the related objects of the target object. For example, the original sentence is \textit{``Man standing''}. We transform the sentence to \textit{``Man standing left to the cat''} to add some constraints to this object by adding a related object. We use COCO annotations to check the existence of the related object, as in the methods above. If the original sentence has two or more nouns, we use the first noun as the target object and change the second to a different, non-exist category name.

\noindent
\textbf{Visual Example.} Fig.~\ref{fig:benchmark} shows a visual example of the five generation methods. In summary, methods (1) and (2) are sentence-level generation approaches, and methods (3), (4), and (5) are word-level generation approaches. They separately focus on the target, attribute, and related object words. These methods can generate many negative sentences with diversity. The final step in creating our validation set is to filter through the data by training a baseline model~\cite{LAVT} to select more challenging negative sentences. We repeat this process twice and obtain the final R-RIS validation sets that are sufficiently difficult. Tab.~\ref{tab:data_statistics} summarizes the statistics of the datasets.

\noindent
\textbf{Proposed R-RIS Datasets.} We build our benchmark based on existing RIS datasets. Specifically, we expand RefCOCO~\cite{RefCOCO}, RefCOCO+~\cite{RefCOCOg}, and RefCOCOg~\cite{RefCOCOg2} to form R-RefCOCO, R-RefCOCO+, and R-RefCOCOg. As negative sentences are not limited in number (non-existing objects can be anything), we generate negative sentences as many as possible. In detail, for each reference, we generate \textit{ten} negative sentences by default.

\begin{figure}[!t]
	\centering
	\includegraphics[width=0.85\linewidth]{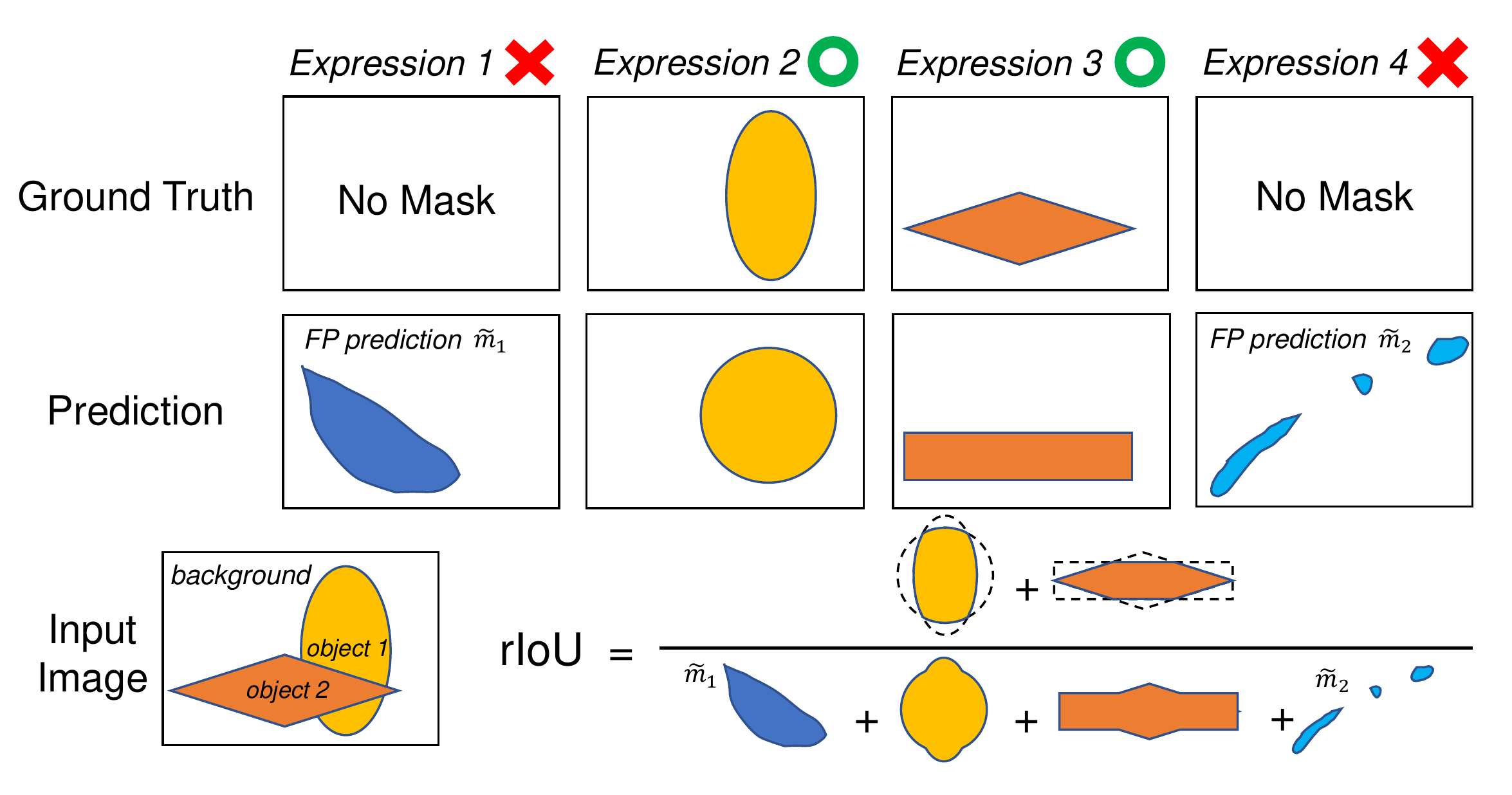}
	\caption{ Illustration of rIoU metric. For positive inputs, both intersection and union terms contribute to the rIoU value. For negative inputs, only the predicted mask contributes to the union term, thus reducing the rIoU value.}
	\label{fig:new-rIoU}
\end{figure}

\noindent
\textbf{New Metrics.} 
%
The existing RIS setting adopts Intersection over Union (IoU) metrics such as mean IoU (mIoU) and overall IoU (oIoU) to evaluate the pixel-level performance. However, they are inadequate in handling negative inputs. Therefore, it is necessary to establish appropriate metrics to validate the R-RIS model's performance.

We propose robust Intersection over Union (rIoU) to consider the negative input jointly. In both RIS and R-RIS settings, data are sampled by reference. Each reference consists of one image, several positive referring expressions, and multiple negative inputs (take R-RIS for examples). 
Supposing the $i$-th reference $\mathcal{R}_i$ has $p_i$ positive sentences and $n_i$ negative sentences, an R-RIS model predicts masks for these language inputs. It gets $\hat{\mathbf{M}}_i = \{\hat{m}^i_1, \hat{m}^i_2, ..., \hat{m}^i_{p_i}, \tilde{m}^i_1, \tilde{m}^i_2..., \tilde{m}^i_{n_i}\}$. There are ground-truth masks $\mathbf{M}_i = \{m^i_1, m^i_2, ..., m^i_{p_i}\}$ for positive inputs. rIoU is calculated as follows:

\begin{equation}
\text{rIoU} = \frac{1}{|\mathcal{R}|} \sum_{i=1}^{|\mathcal{R}|} \frac{\sum_{j=1}^{p_i} |\hat{m}^i_j \cap m^i_j|}{\sum_{j=1}^{p_i} |\hat{m}^i_j \cup m^i_j| + \sum_{k=1}^{n_i} |\tilde{m}^i_k|}
\end{equation}

where $|\mathcal{R}|$ represents number of references in the validation set, and $|m|$ denotes the number of pixels in mask $m$. 
$\cap$ and $\cup$ refer to the intersection and union operations. 
However, the denominator includes the mask predicted by the model for negative inputs, which can also be expressed as $|\tilde{m}^i_k \cup \emptyset|$ since the ground truth for negative inputs is always $\emptyset$. 
This aspect treats the mask predicted for negative inputs as the union component, thereby penalizing the model for incorrect outputs for negative inputs. Figure~\ref{fig:new-rIoU} illustrates the rIoU metric.

Despite rIoU being a metric for R-RIS, there is a need for a more precise assessment of the model's discriminative ability on negative inputs. In most cases, an R-RIS model is expected to generate small masks for negative inputs and an exact 0-pixel mask, indicating no mask. Therefore, we adopt an instance-level metric, mean Robust Recall (mRR), to evaluate the R-RIS model's performance. 

For the input reference $\mathcal{R}_i$, the Robust Recall is computed as $\mathrm{RR}_i = \frac{1}{n_i} \sum^{n_i}_{k=1} \mathbb{I} (|\tilde{m}^i_k| = 0)$, where $\mathbb{I}(\cdot)$ is the indicator function, which equals 1 if the input is an exact 0-pixel mask and 0 otherwise. mRR is then defined as the mean RR across all references in the validation set, as shown below:

\begin{equation}
\mathrm{mRR} = \frac{1}{|\mathcal{R}|} \sum^{|\mathcal{R}|}_{i=1} \mathrm{RR}_i\ .
\end{equation}

Since R-RIS is still a pixel-level prediction task, we use rIoU as the main metric. Moreover, only using mRR can not well judge the extreme cases where the R-RIS models treat all inputs as negative examples.
\begin{figure*}[!h]
	\centering
	\includegraphics[width=1.0\linewidth]{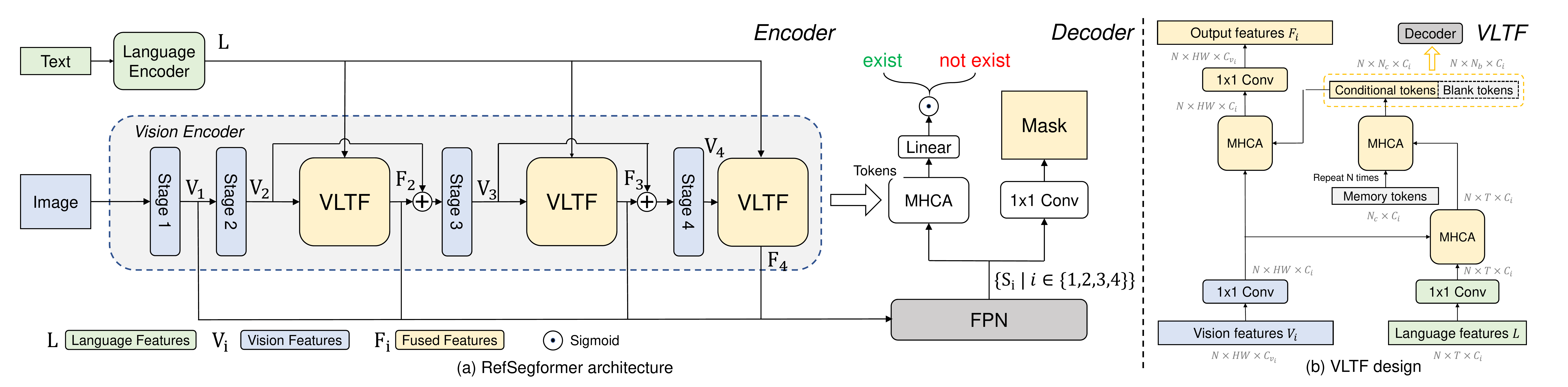}
	\caption{ (a) The architecture of RefSegformer. It contains a vision encoder, a language encoder, an encoder fusion meta-framework that contains our proposed VLTF module for connecting language features and vision features, and an FPN-like decoder with a binary classification head. (b) VLTF design. It consists of three multi-head cross-attention modules (MHCAs). Blank tokens are concatenated with conditional tokens to avoid conditional tokens over-fitting on negative text inputs.}
	\label{fig:method_overview}
\end{figure*}

\section{Method}

In this section, we provide a detailed account of the RefSegformer. 
Initially, we introduce a two-stage baseline for the R-RIS task and subsequently delve into the architecture of RefSegformer. The proposed model incorporates a novel module called Vision-Language Token Fusion (VLTF). 
To better handle the R-RIS task, our framework also introduces a negative token modeling technique, which offers a simple yet effective baseline for R-RIS.

\subsection{RefSegformer Architecture}
\label{sec:method_baseline}

\noindent
\textbf{Two-Stage Baseline.}
To tackle the R-RIS task, it is intuitive to develop a two-stage model that integrates a standard RIS model and an extra binary classifier to verify whether the referring expression is correct. The model generates a mask only when the classification result is true. Drawing inspiration from prior studies~\cite{EFN, LAVT}, we establish our RefSegformer from a two-stage baseline by fusing vision and language features early in the encoding stage.

\noindent
\textbf{Encoder Fusion Framework.}
Given input containing a pair of an image and a referring expression, we leverage a language encoder~\cite{BERT} to extract the features of the expression, denoted as $L \in \mathbb{R}^{T \times C_l}$, where $T$ is the number of words, and $C_l$ is the number of channels. These language features are inserted into the vision encoder and fused with vision features through our proposed Vision-Language Token Fusion module (VLTF) between each vision stage. Specifically, we adopt Swin Transformer~\cite{swin} as our vision encoder. 
The four stages of the vision encoder are denoted as $\{v_i | i \in \{1, 2, 3, 4\}\}$, while VLTF modules are $\{f_i | i \in \{2, 3, 4\}\}$.
The whole encoding process can be formulated as follows:
\begin{equation}
\left\{
\begin{aligned}
V_1 &= v_1(I) \\
V_2 &= v_2(V_1) \\
F_i &= f_i(V_i, L), i \in \{2, 3, 4\} \\
V_i &= V_{i-1} + \sigma(F_i), i \in \{3, 4\},
\end{aligned}
\right.
\end{equation}
where $I$ denotes the input image, $\{V_i | i \in \{1, 2, 3, 4\}\}$ represents vision features from four stages in Swin Transformer, and $\{F_i | i \in \{2, 3, 4\}\}$ denotes the fused multi-modal features. $\sigma$ is a normalization function. Following each stage in Swin Transformer, except for stage 1, a VLTF module combines vision and language features, fusing them to obtain a multi-modal feature $F_i$. Subsequently, the multi-modal feature is fed as input to the next stage. Ultimately, we obtain three hierarchical multi-modal features $\{F_i | i \in {2, 3, 4}\}$ and one vision feature $V_1$ with high resolution, which we feed into an FPN-like decoder to extract the hierarchical mask features. We utilize a stack of $1 \times 1$ convolution layers to produce the final mask.

\noindent
\textbf{FPN-like Decoder.}
To generate the final masks, we use an FPN-like decoder that takes the hierarchical features of $V_1$ and $\{F_i | i \in \{2, 3, 4\}\}$ as input. The decoder outputs $\{S_i | i \in \{1, 2, 3, 4\}\}$, which are then passed to $1 \times 1$ convolutional layers, separately, to obtain coarse-to-fine segmentation masks.
\begin{equation}
\left\{
\begin{aligned}
S &= \phi(V_1, F_2, F_3, F_4) \\
M &= \rho(S_i), i \in \{1, 2, 3, 4\}, \\
\end{aligned}
\right.
\end{equation}
where $S = \{S_i | i \in \{1, 2, 3, 4\}\}\}$ represents the segmentation features, $M = \{M_i | i \in \{1, 2, 3, 4\}\}\}$ represents the hierarchical output masks, $\phi$ denotes the FPN-like decoder, and $\rho$ denotes the $1 \times 1$ convolutions that project the segmentation features into two-dimensional mask score maps. Finally, we calculate the cross-entropy loss, shown as follows:
\begin{equation}
L_s = \text{CELoss}(y, M_1) + \lambda \cdot \sum_{i=2}^4\text{CELoss}(y, M_i),
\label{loss:ce_baseline}
\end{equation}
where $\lambda$ is a hyperparameter between 0 and 1 that reduces the impact of coarse-grained masks.

\noindent
\textbf{Binary Classification Head.}
We add a binary classification head upon the outputs of the FPN-like decoder and the tokens from a VLTF module. Specifically, we utilize the mask feature $M_1$ with the highest resolution to interact with the conditional and blank tokens from the last VLTF using an MHCA. Subsequently, we feed the output into a linear layer and obtain the predicted probability of the existence:

\begin{equation}
    \hat{e} = \mu(\text{MHCA}(M_1, T_4, T_4)),
\end{equation}

where $T_4$ refers to the tokens from the last VLTF module, and $\mu$ denotes a linear layer. By default, $M_1$ is used as the query, while $T_4$ serves as the key and value. In our experiments, we test the opposite approach and find that using the high-resolution features as the query yield superior results.

An additional existing loss is applied to optimize the binary classification head.
\begin{equation}
L_e = \-{(e\log(\hat{e}) + (1 - e)\log(1 - \hat{e}))},
\end{equation}
where $e$ is the ground truth about the existence of the referred object. Finally, the complete loss function in our training procedure for the R-RIS task is:
\begin{equation}
L = L_s + \gamma \cdot L_e,
\end{equation}
where $\gamma$ is a hyperparameter balancing segmentation loss and existing loss. We set $\gamma$ to 1.0 in our experiments.

\subsection{Vision-Language Token Fusion Module}

As shown in Fig~\ref{fig:method_overview}(b), the Vision-Language Token Fusion module (VLTF) is the key component for RefSegformer. 
It contains several attention operations to fuse vision and language features via \textit{learnable memory tokens}.

Given an input vision feature $V_i \in \mathbb{R}^{N\times HW\times C_{v_i}}$ of stage $i$ and language feature $L \in \mathbb{R}^{N\times T\times C_l}$, we first transform them into a common dimension $C_i$ using two separate $1\times 1$ convolution layers. The transformed vision and language features are then passed through an MHCA module, with the language features serving as the query and the vision features as key and value. This process produces a vision-aware language-shaped feature. We then introduce a set of randomly initialized vectors known as \textit{memory tokens} to perform the second MHCA operation. This results in the generation of \textit{conditional tokens}, which contain multi-modal information. 
The conditional tokens are further processed using the third MHCA module and the transformed vision features. At this time, the transformed vision features serve as the query, projecting the language information into the vision features. Finally, through a $1 \times 1$ convolution, the output feature of VLTF is of the same shape as the input vision features but is now fused with the information of language, resulting in a multi-modal fused feature.
Unlike previous works~\cite{EFN,LAVT} that adopt pixel-wised cross-modal attention, our framework adopts memory tokens to dynamically select relevant language information, which is flexible to extend into the R-RIS setting. 

\subsection{Extension to Robust Referring Segmentation}
\label{sec:method_robust}

\noindent
\textbf{Blank Tokens Design.}
In the second MHCA of VLTF, it takes randomly initialized memory tokens and vision-aware language-shaped features as input and outputs a set of vectors $T_{i_c} \in \mathbb{R}^{N\times K_c\times C_i}$. We call them conditional tokens. 

To further handle the R-RIS task, we a new set of blank tokens, represented as $T_{i_b} \in \mathbb{R}^{N\times K_b\times C_i}$. These blank tokens are concatenated with the set of conditional tokens $T_{i_c}$, as depicted in Fig~\ref{fig:method_overview}(b). Notably, the blank tokens are randomly initialized and are not fused with the language features. When fed to the third MHCA, the blank tokens interact with vision features and are expected to attend to linguistically unrelated regions. This approach effectively decouples the effectiveness of conditional and blank tokens, enabling the former to learn the alignments between vision and language, while the latter serves as an unrelated linguistic learner. Thus expanding our model to the R-RIS task. 

\noindent
\textbf{R-RIS Training Pipeline.}
We train the R-RIS models by adding negative sentences into the training set. The model is to learn to distinguish them by setting the ground-truth segmentation mask of negative sentences to all-0. We set the ratio between positive and negative sentences to 1:1. Our proposed training pipeline requires no changes to most RIS models' architecture or loss functions, making it a straightforward and practical solution for R-RIS model training.

\section{Experiment}

\begin{table}[t]
   \centering
   \scalebox{0.52}{
   \setlength{\tabcolsep}{3.6mm}{\begin{tabular}{l|c|c|c|c|c|c|c|c|c}
      \toprule[1pt]
      \multirow{2}{*}{Method} & \multicolumn{3}{c|}{R-RefCOCO}  & \multicolumn{3}{c|}{R-RefCOCO+} & \multicolumn{3}{c}{R-RefCOCOg} \\
      \cline{2-10}
                        & mIoU & mRR & rIoU & mIoU & mRR & rIoU & mIoU & mRR & rIoU  \\
      \hline
      CRIS~\cite{CRIS}  & 43.58 & 76.62 & 29.01 & 32.13 & 72.67 & 21.42 & 27.82	& 74.47 &	14.60 \\
      EFN~\cite{EFN}    & 58.33  & 64.64 & 32.53 & 37.74  & 77.12 & 24.24 & 32.53 & 75.33 & 19.44 \\
      VLT~\cite{VLT}    &  61.66  & 63.36 & 34.05 & 50.15 & 75.37 & 34.19 & 49.67 & 67.31 & 31.64\\
      LAVT~\cite{LAVT}   & 69.59 & 58.25 & 36.20 & 56.99 & 73.45 & 36.98 & 59.52 &  61.60 & 34.91 \\
      LAVT+~\cite{LAVT}     & 54.70 & 82.39 & 40.11 & 45.99 & 86.35 & 39.71 & 47.22 & 81.45 & 35.46 \\
      \hline
      RefSegformer & 68.78 & 73.73 & \textbf{46.08} & 55.82 & 81.23 & \textbf{42.14} & 54.99 &  71.31 & \textbf{37.65} \\ 
      \bottomrule[1pt]
   \end{tabular}}}
   \caption{ Comparison to other state-of-the-art models for the R-RIS task.}
   \label{tab:result_R-RIS}
\end{table}

\subsection{Dataset and Settings}

\noindent
\textbf{Dataset and Metrics} We build our benchmark based on three standard RIS datasets, RefCOCO \cite{RefCOCO}, RefCOCO+ \cite{RefCOCO}, and RefCOCOg \cite{RefCOCOg, RefCOCOg2}, formulating R-RefCOCO, R-RefCOCO+, and R-RefCOCOg. We abandon the words about absolute positions when generating R-RefCOCO+ following~\cite{RefCOCO}. We conduct our robust benchmark on the UNC/UMD partitions and target only the validation set. The three robust validation sets can be called as `rval' set if referred to in future works. For our proposed R-RIS setting, we adopt robust Intersection-over-Union (rIoU) and mean Robust Recall (mRR) as the main metrics. In addition, we report standard metrics, including mean intersection-over-union (mIoU), overall intersection-over-union (oIoU), and precision at the different thresholds. Precision means the percentage of test samples that IoU number is higher than the threshold.

\noindent
\textbf{Implementation Details.}
We use PyTorch to implement our model. The language encoder is an officially pre-trained BERT model~\cite{BERT}. For the vision encoder, we adopt Swin Transformer~\cite{swin} as the architecture and initialize it with the pre-trained weights on ImageNet~\cite{deng2009imagenet}. Other modules, such as VLTF and the segmentation decoder, are trained from scratch. We set the hyperparameter $\lambda$ to 0.4. The numbers of conditional and blank tokens are set to 20 and 10, respectively. The maximum length of input referring expressions is set to 20 for all datasets. We use the AdamW optimizer~\cite{ADAMW} with a weight decay of 0.02. The learning rate is initialed as 3e-5 and scheduled by polynomial learning rate decay with a power of 0.9. Following~\cite{ReSTR,LAVT}, input images are resized to 480x480 without any other data augmentations. All the models are trained for 50 epochs with batch size 64 on 8 Nvidia RTX-3090 GPUs.

\subsection{Benchmark Results}

\noindent
\textbf{Results on R-RIS datasets.} We benchmark five different methods on our proposed datasets using the same training setting for a fair comparison. 
Tab.~\ref{tab:result_R-RIS} presents the results of the R-RIS task for RefSegFormer and these models, including CRIS~\cite{CRIS}, EFN~\cite{EFN}, VLT~\cite{VLT}, LAVT~\cite{LAVT} and LAVT+~\cite{LAVT}. 
We denote LAVT+ as LAVT~\cite{LAVT} combined with our binary classification head, which is a strong baseline for reference.
All models are trained using the pipeline described in Sec.~\ref{sec:method_robust}. RefSegFormer outperforms all baseline models in our proposed rIoU metric. Furthermore, compared to most baseline models, RefSegFormer achieves a better balance between mIoU and the mean Robust Recall mRR. 
Although LAVT+ performs well in mRR, it still has much lower rIoU. 
This is because the model cannot effectively disentangle between positive and negative inputs, leading to a trivial output of mostly negative. 
That means all examples may be negative ones. We will provide more examples in the appendix. 
In contrast, our method can achieve much better results (6\% rIoU better than LAVT+) to balance both positive and negative inputs.

\noindent
\textbf{Results on RIS datasets.} We further present the performance of RefSegformer in the regular RIS task. As depicted in Tab.~\ref{tab:result_RIS}, although RefSegformer is primarily designed to excel in the proposed R-RIS task, without further extra design, it achieves strong performance in regular RIS as well. 
Notably, our method achieves eight out of nine best results on these datasets without specific modifications. 
RefSegformer outperforms the previous state-of-the-art method, LAVT~\cite{LAVT}. The most notable improvement of 1.38 is observed in the val split of RefCOCO+. An average improvement of 0.43 is achieved across all nine validation sets. This result underlines the exceptional generalization ability of RefSegformer.

\begin{table*}[!t]
   \centering
   \scalebox{0.75}{
   \setlength{\tabcolsep}{2.5mm}{\begin{tabular}{l|l|c|c|c|c|c|c|c|c|c}
      \toprule[1pt]
      \multirow{2}{*}{Method} &
      Language &
      \multicolumn{3}{c|}{RefCOCO}  & \multicolumn{3}{c|}{RefCOCO+} & \multicolumn{3}{c}{RefCOCOg} \\
      \cline{3-11}
                                    & Model & val   & test A & test B & val & test A & test B & val (U) & test (U) & val (G)  \\
      \hline
      MAttNet~\cite{yu2018mattnet}  & Bi-LSTM & 56.51 & 62.37 & 51.70 & 46.67 & 52.39 & 40.08 & 47.64 & 48.61 & -     \\
      CMSA~\cite{ye2019cross}       & None & 58.32 & 60.61 & 55.09 & 43.76 & 47.60 & 37.89 & -     & -     & 39.98 \\
      CAC~\cite{chen2019referring}        & Bi-LSTM & 58.90 & 61.77 & 53.81 & -     & -     & -       & 46.37 & 46.95 & 44.32 \\
      LSCM~\cite{hui2020linguistic} & LSTM & 61.47 & 64.99 & 59.55 & 49.34 & 53.12 & 43.50 & -     & -     & 48.05 \\
      CMPC+~\cite{CMPC}      & LSTM & 62.47 & 65.08 & 60.82 & 50.25 & 54.04 & 43.47 & -     & -     & 49.89 \\
      MCN~\cite{MCN}       & Bi-GRU & 62.44 & 64.20 & 59.71 & 50.62 & 54.99 & 44.69 & 49.22 & 49.40 & -     \\
      EFN~\cite{EFN}                & Bi-GRU & 62.76 & 65.69 & 59.67 & 51.50 & 55.24 & 43.01 & - & - & 51.93     \\
      BUSNet~\cite{BUSNet}          & Self-Att & 63.27 & 66.41 & 61.39 & 51.76 & 56.87 & 44.13 & - & - & 50.56 \\
      CGAN~\cite{CGAN}    & Bi-GRU & 64.86 & 68.04 & 62.07 & 51.03 & 55.51 & 44.06 & 51.01 & 51.69 & 46.54 \\
      ISFP~\cite{liu2022instance}&Bi-GRU&{65.19} & {68.45} & {62.73} & {52.70} & {56.77} & {46.39} & {52.67} & {53.00} & 50.08 \\
      LTS~\cite{LTS}   & Bi-GRU & 65.43 & 67.76 & 63.08 & 54.21 & 58.32 & 48.02 & 54.40 & 54.25 & - \\
      VLT~\cite{VLT}      & Bi-GRU & 65.65 & 68.29 & 62.73 & 55.50 & 59.20 & 49.36 & 52.99 & 56.65 & 49.76 \\
      ReSTR~\cite{ReSTR}    & Transformer & 67.22 & 69.30 & 64.45 & 55.78 & 60.44 & 48.27 & 54.48 & - & - \\
      CRIS~\cite{CRIS} & Transformer & 70.47 & 73.18 & 66.10 & 62.27 & 68.08 & 53.68 & 59.87 & 60.36 & - \\
      LAVT\cite{LAVT} & BERT & {72.73} & \underline{75.82} & {68.79} & {62.14} & {68.38} & {55.10} & {61.24} & {62.09} & \textbf{60.50} \\
      \hline
      RefSegformer & BERT & \underline{73.22} & 75.64 & \underline{70.09} & \underline{63.50} & \underline{68.69} & \underline{55.44} & \textbf{62.56} & \textbf{63.07} & \underline{58.48} \\
      RefSegformer* & BERT & \textbf{76.92} & \textbf{79.31} & \textbf{74.25} & \textbf{63.69} & \textbf{69.24} & \textbf{55.91} & \underline{61.29} & \underline{62.42} & 56.47 \\
      \bottomrule[1pt]
   \end{tabular}}}
   \caption{ Comparison with state-of-the-art methods in RIS task U: The UMD partition. G: The Google partition. We refer to the language model of each reference method. * means with text prompt. }
   \label{tab:result_RIS}
\end{table*}

\subsection{Ablation Study and Analysis}

\begin{table}[!t]
   \centering
   \scalebox{0.75}{
   \setlength{\tabcolsep}{2mm}{\begin{tabular}{c|c|c|c|c|c}
      \toprule[1pt]
      \multirow{2}{*}{\# Blank Tokens} & Binary & \multirow{2}{*}{mIoU} & \multirow{2}{*}{oIoU} & \multirow{2}{*}{mRR} & \multirow{2}{*}{rIoU} \\
      & Head & & & & \\
      \hline
      \multirow{3}{*}{0} & \textit{None} & 69.52 & 70.25 & 58.50 & 31.67 \\ 
      & T & 69.57 & 70.29 & 68.63 & 38.10 \\
      & V & 68.96 & 69.72 & 70.38 & \textbf{42.01} \\
      \hline
      \multirow{3}{*}{5} & \textit{None} & 70.14 & 70.66 & 58.46 & 35.40 \\ 
      & T & 69.01 & 69.93 & 69.57 & 41.81 \\
      & V & 68.93 & 69.35 & 70.86 & \textbf{45.65} \\
      \hline
      \multirow{3}{*}{10} & \textit{None} & 70.50 & 70.63 & 58.80 & 38.78 \\ 
      & T & 69.01 & 69.26 & 68.40 & 44.12 \\
      & V & 68.78 & 69.78 & 73.73 & \textbf{46.08} \\
      \bottomrule[1pt]
   \end{tabular}}}
   \caption{ Ablation studies on the R-RefCOCO dataset for R-RIS. In the second column, ``None'' means without binary classification head, ``T'' means VLTF tokens as the query, and ``V'' means vision features as the query.}
   \label{tab:rRIS_ablation}
\end{table}

\noindent
\textbf{Blank Tokens Design.}
We investigate the impact of blank tokens on the performance of VLTF. We conduct experiments by training our model with 0, 5, and 10 blank tokens and report the results in Tab.~\ref{tab:rRIS_ablation}. Compared to the no blank tokens setting, the default setting of RefSegformer, which uses 10 blank tokens, yields an improvement of 4.07 total points in rIoU. Increasing the number of blank tokens from 0 to 5 also improves rIoU by 3.54. and increasing it to 10 results in the most significant effect.

\noindent
\textbf{Binary Classification Head.}
We discuss the importance of the binary classification head in RefSegformer for the R-RIS task. As shown in Tab.~\ref{tab:result_R-RIS}, when \textit{without} this explicit exist-or-not classification head, the performance experiences a significant drop with an average decay point around 6. Once added, the model's performance significantly improves, regardless of whether we input tokens or vision features as the query. This result underscores the importance of explicitly predicting whether the referred object occurs in the image. Moreover, our experiments demonstrate that vision features outperform tokens in all settings for choosing MHCA input queries. This result is because a multi-modal fused feature map contains richer information than a set of embeddings.

\begin{table}[!t]
   \centering
   \footnotesize
   \scalebox{0.90}{
   \setlength{\tabcolsep}{2.5mm}{\begin{tabular}{c|c|c|c|c|c|c}
      \toprule[1pt]
      \#VLTFs & mIoU & oIoU & mRR & rIoU & GFLOPs & Params \\
      \hline
      1 & 65.35 & 66.30 & 68.48 & 39.40 & 127.86 & 229.19M \\
      2 & 68.33 & 68.46 & 72.90 & 43.33 & 130.40 & 233.27M \\
      3 & 68.78 & 69.78 & 73.73 & 46.08 & 132.86 & 234.39M \\
      \bottomrule[1pt]
   \end{tabular}}}
   \caption{ Ablation study on the effectiveness and computational cost of VLTFs.}
   \label{tab:VLTF_ablation}
\end{table}

\noindent
\textbf{Effectiveness of VLTF.}
We evaluate the effectiveness of the VLTF module in RefSegformer. We vary the number of VLTF modules and show the results in Tab. \ref{tab:VLTF_ablation}. Our experiments demonstrate that the model's performance decreases as we use fewer VLTFs, indicating that the proposed multi-modal fusion network on multi-stage is essential and effective for the R-RIS task. Specifically, the performance declines from 46.08 to 39.40 in terms of rIoU when using only one VLTF module, while the performance drops from 46.08 to 43.33 when using two VLTF modules.

\begin{figure}[!t]
	\centering
	\includegraphics[width=0.60\linewidth]{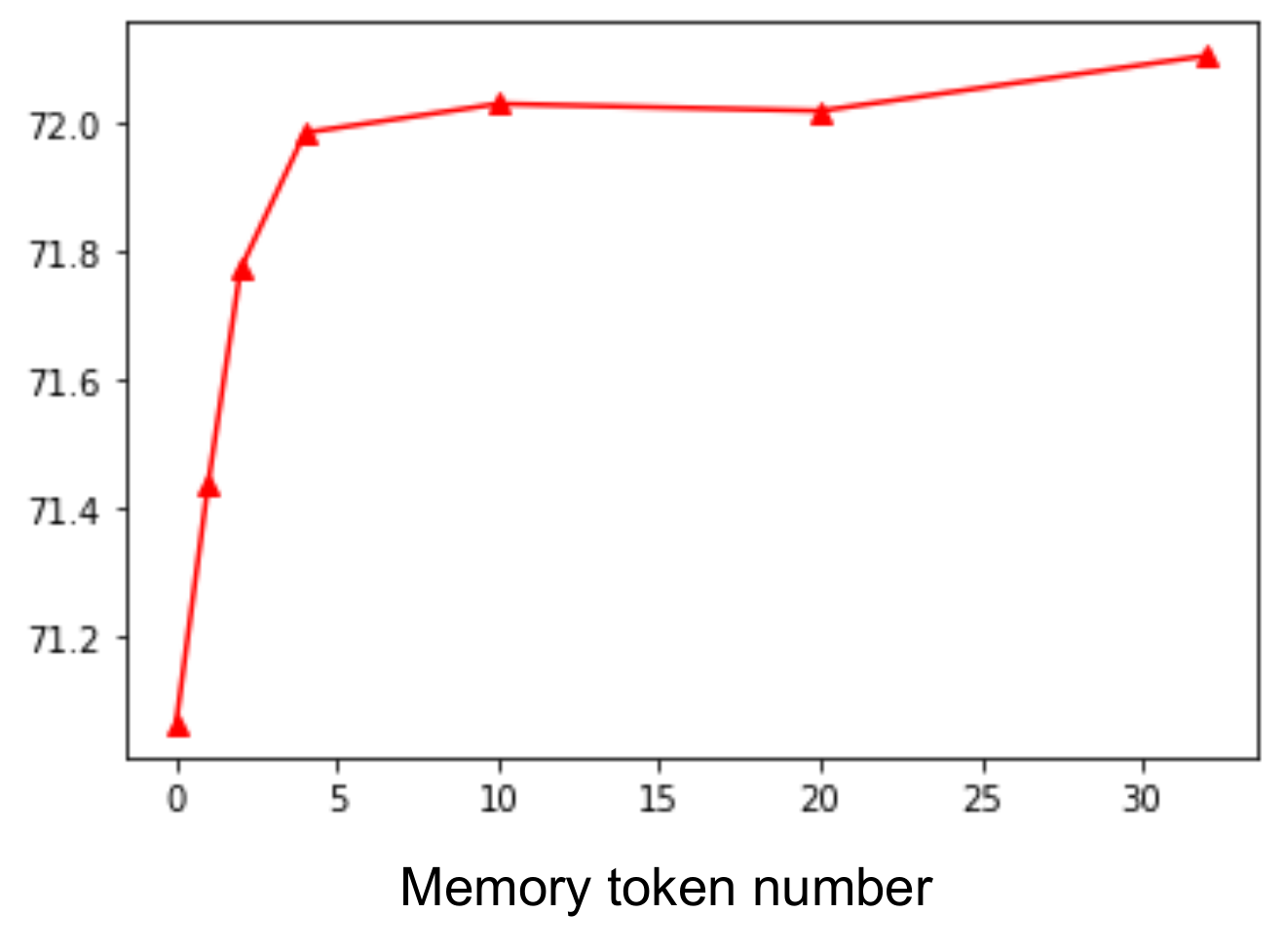}
	\caption{ oIoU gains by increasing memory tokens number $N_c$.}
	\label{fig:visual_compare}
\end{figure}

\noindent
\textbf{Number of Memory Tokens.} We explore the number of memory tokens for the RIS dataset. When increasing the number of memory tokens in VLTF, the oIoU metrics increase initially and then become saturated. This is because the relevant language semantics are limited in the RIS dataset. We choose $K_c = 20$ as the default setting.

\begin{table}[!t]
   \centering
   \footnotesize
   \scalebox{0.90}{
   \setlength{\tabcolsep}{3.3mm}{\begin{tabular}{c|c|c|c|c|c}
      \toprule[1pt]
      Deformable & Text & \multirow{2}{*}{P@0.5} & \multirow{2}{*}{P@0.7} & \multirow{2}{*}{P@0.9} & \multirow{2}{*}{oIoU} \\
      FPN & Prompt & & & & \\
      \hline
      & & 90.71 & 84.42 & 68.76 & 72.02 \\
      \checkmark & & 91.66 & 87.05 & 71.40 & 73.22 \\
      \checkmark & \checkmark & 87.84 & 76.73 & 38.57 & 76.92 \\
      \bottomrule[1pt]
   \end{tabular}}}
   \caption{ Ablations about components of RefSegFormer for RIS task. Evaluated in val split, RefCOCO dataset.}
   \label{tab:RIS_ablation}
\end{table}

\noindent
\textbf{Deformable FPN and Text Prompt.}
To generate more precise results, we adopt the recently proposed deformable pixel decoder~\cite{deformable-DETR} to decode the multi-modal features $F$ dynamically. By comparing with the plain FPN architecture, we improve 1.2\% in oIoU, as demonstrated in Tab.~\ref{tab:RIS_ablation}. Moreover, we utilize prompting methods by concatenating multiple referring expressions into one sentence~\cite{liu2021pre}. For example, ``woman on the right'', and ``woman in black'' are concatenated as ``woman on the right woman in black''. As shown in Tab.~\ref{tab:RIS_ablation}, text prompt leads to a significant increase of 3.7 in the oIoU value, albeit with a decreased precision. This is because adding more text can improve recognition ability while decreasing localization ability since more localization text noises become the inputs.

\noindent
\textbf{GFLOPs and Parameter Analysis.}
To analyze the computational cost of RefSegformer, we evaluate the GFLOPs and parameters at varying numbers of VLQFs, as presented in Tab. \ref{tab:VLTF_ablation}. We observe that as the number of VLQFs increases from 1 to 3, the GFLOPs show a slight increment of 5\%. Similarly, the parameters increase by 4.2M. Nevertheless, these increases are negligible compared to the overall computational cost of the model.

\begin{figure}[!h]
	\centering
	\includegraphics[width=1.0\linewidth]{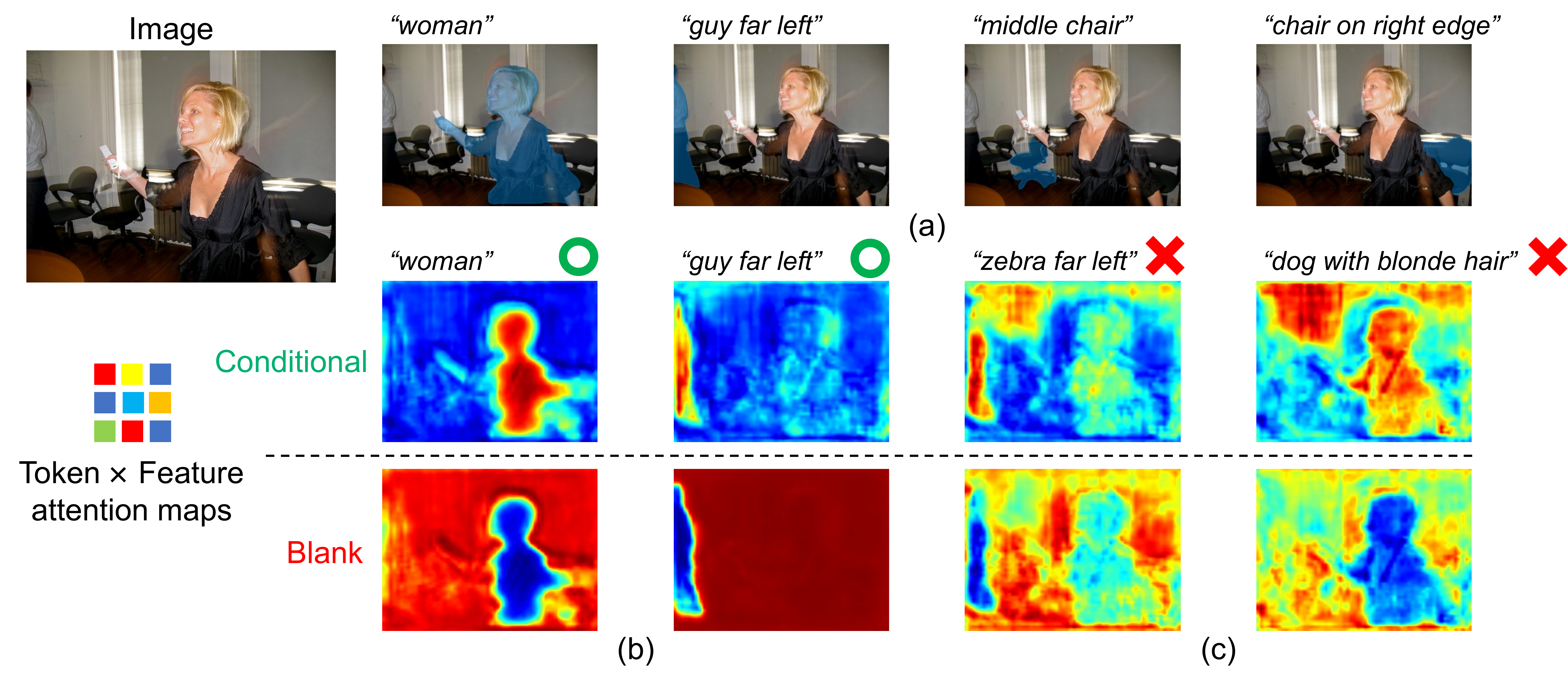}
	\caption{ Image-token attention visualization. (a) Prediction of positive and negative inputs. (b) Attention maps between conditional/blank tokens and the visual features for positive sentences. (c) Attention maps for negative sentences.}
	\label{fig:visualization_analysis}
\end{figure}

\begin{figure}[!t]
	\centering
	\includegraphics[width=1.0\linewidth]{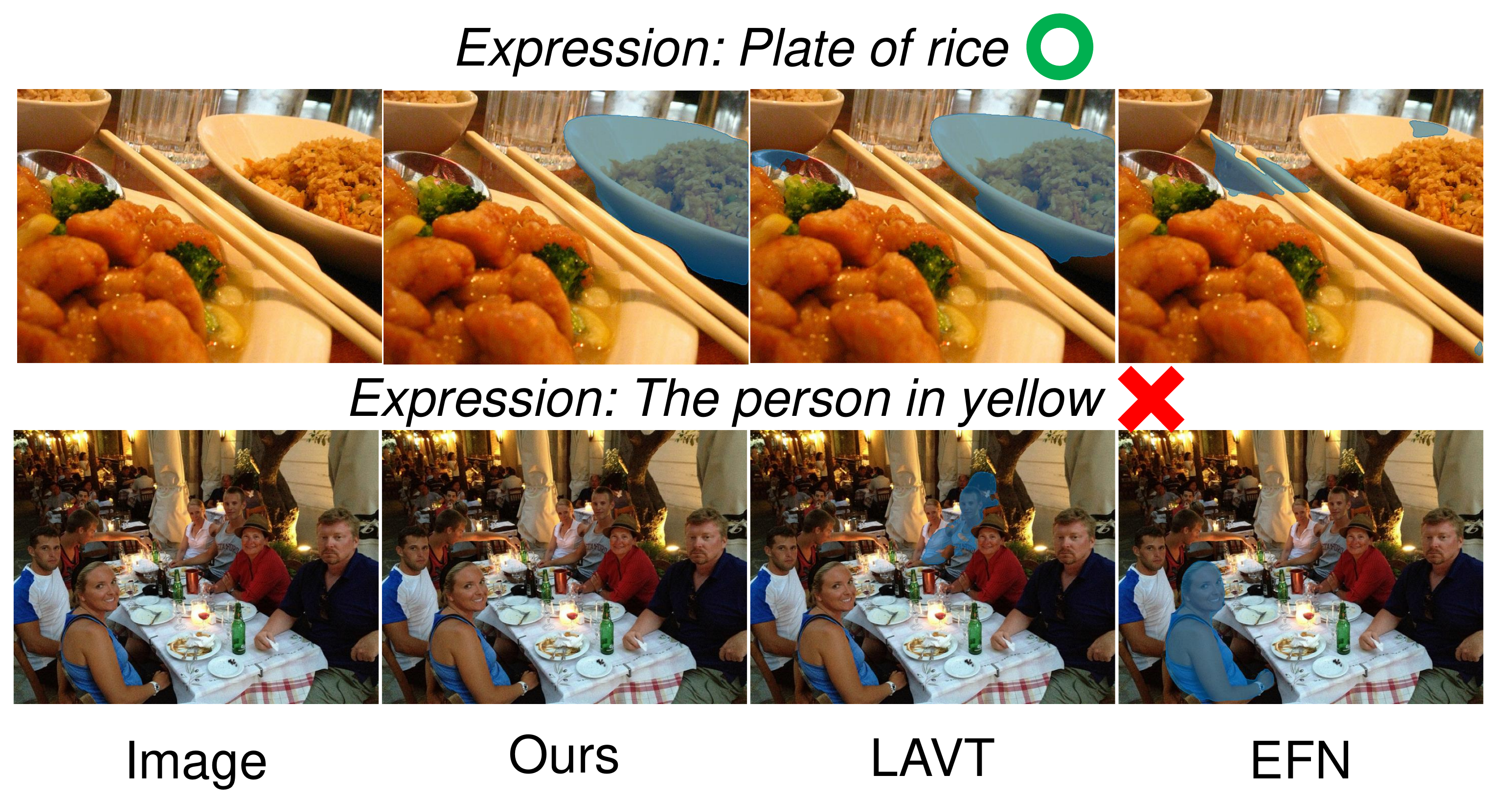}
	\caption{ Visual Comparison with previous methods~\cite{LAVT,EFN}. Our model shows both accuracy and robustness for the R-RIS task.}
	\label{fig:visual_compare}
\end{figure}

\noindent
\textbf{Visualization Analysis.} Fig.~\ref{fig:visualization_analysis} (a) shows RefSegformer can perform correct segmentation results given various language expressions. In Fig.~\ref{fig:visualization_analysis} (b) and (c), we visualize the attention maps between conditional/blank tokens and the visual features. We observe that blank tokens and conditional tokens focus on the opposite areas, demonstrating that the blank tokens learn to attend to linguistic-unrelated areas. 

In Fig~\ref{fig:visual_compare}, we compare RefSegformer with state-of-the-art baselines~\cite{EFN, LAVT}. RefSegformer outputs correctly for both positive inputs (top row) and negative sentences (bottom row). More visible results, including failure cases study, can be found in the supplementary material. 

\section{Conclusion}
We propose a novel task called Robust Referring Image Segmentation (R-RIS), which takes both positive and negative text inputs into consideration to achieve a robust and explainable machine learning approach. To build robust referring image segmentation datasets, \textit{five} different ways of selecting negative texts have been presented. We also introduce new metrics, named rIoU and mRR, that effectively evaluate the R-RIS models. In order to solve the R-RIS problem, we propose a transformer-based baseline, RefSegformer. It connects the image and text features through learnable tokens, and the token-based approach can be easily extended to R-RIS by adding blank tokens. The proposed RefSegformer achieves state-of-the-art results on both RIS datasets and R-RIS datasets, thereby serving as a new baseline for future research in this area. We believe these works will inspire further studies in this field.

\noindent
\textbf{Future Works.} Two potential research directions on R-RIS can be: 1, train only one model for both R-RIS and RIS settings. 2, a better model to balance mRR and rIoU. 

\noindent
\textbf{Acknowledgement.} This work is supported by the National Key Research and Development Program of China (No.2020YFB2103402).
We also thank Xiang Li from CMU for the discussion of the tasks and metric design.


\appendix

\section{Appendix}

\noindent
\textbf{Overview} As mentioned in the main paper, we provide the following parts: 1, More building benchmark details. 2, More implementation details. 3, More ablation studies and comparative analysis of the proposed baseline RefSegformer, including more comparison, visual examples, and analysis. 4. Comparison with recent co-current work, R2VOS~\cite{R2VOS}.

\subsection{More Benchmark Details}
\label{sec:more_benchmark_details}

\begin{table}[!t]
   \centering
   \scalebox{0.65}{
   \setlength{\tabcolsep}{2mm}{\begin{tabular}{c|l|c}
      \toprule[1pt]
      Method & Description & Mask Ratio \\
      \hline
      Sentence & Replace the entire sentence with another sentence & 7.93 \\
      Category & Replace the entire sentence with a random category name & 11.37 \\
      Target Obj. & Replace the target object with a random category name & 9.48 \\
      Attribute & Change the attribute words of the target object & 9.64 \\
      Relation Obj. & Change the relation object & 10.26 \\
      \hline
      \textit{Origin} & \textit{No negative sentence} & 10.97 \\
      \bottomrule[1pt]
   \end{tabular}}}
   \caption{Five approaches to generate negative sentences and the test results of a state-of-the-art RIS model~\cite{LAVT}. The mask ratio is the average size of the output masks over the entire image. ``Origin'' means the results on the origin RIS dataset.}
   \label{tab:negative-description}
\end{table}

\noindent
\textbf{RIS Model on Negative Sentences.}
We conduct an experiment that uses a previous model \cite{LAVT} trained on the RIS datasets for the evaluation of negative sentences. We report the average size of the output masks over the size of the whole image. Tab.~\ref{tab:negative-description} shows LAVT outputs masks with an average of around $1/10$ of the image size for the RIS task, originally. When given negative sentences as input, the ratios decrease slightly. For the ``Category'' generation method, the number is even higher. The results demonstrate that a RIS model trained only with positive sentences cannot be directly transferred to the proposed R-RIS task.

\subsection{More Implementation Details}
\label{sec:implementation_details}

\begin{figure}[!t]
	\centering
	\includegraphics[width=1.0\linewidth]{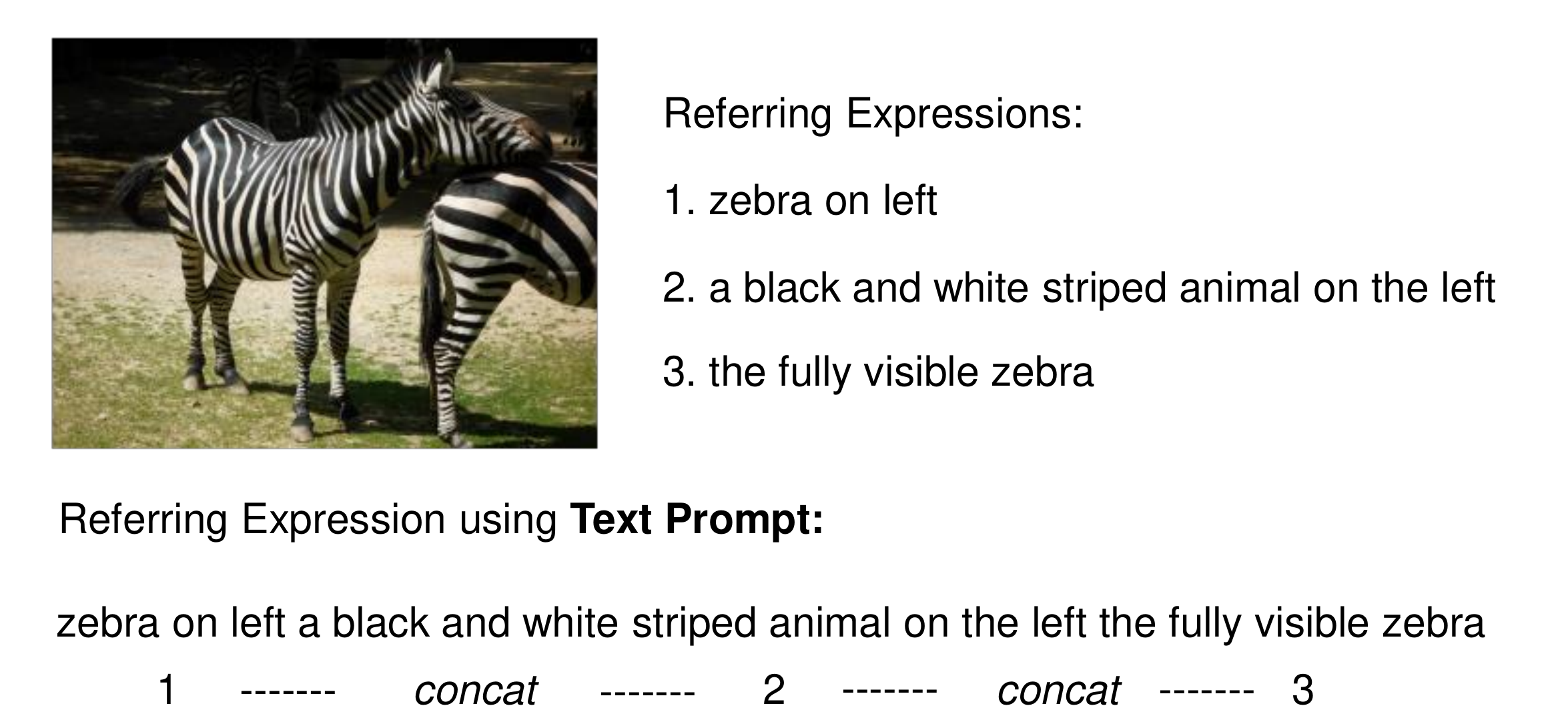}
	\caption{An example of text prompt in evaluation for illustration.}
	\label{fig:sup_text_prompt}
\end{figure}

\noindent
\textbf{Text Prompt.}
In RefCOCO/+/g, there are usually several sentences describing the same object in an image. As shown in Fig. \ref{fig:sup_text_prompt}, a zebra can be described in three different sentences. When evaluating the traditional RIS task, we propose a setting called `with prompt.' The details are also shown in Fig. \ref{fig:sup_text_prompt}. Instead of evaluating the same reference several times with three sentences and averaging the IoU value for the final metric, we alternatively concatenate all three sentences into one \textbf{long} sentence (bottom in Fig. \ref{fig:sup_text_prompt}). Doing so can reduce the validating time because the model only needs to forward once for one reference. At the same time, we observe a result increase in the final oIoU metric (See Sec. 5). That is probably because a concatenated long sentence contains more information than a separate one does individually.

\subsection{More Experimental Results}
\label{sec:more-exp}

\begin{table}[!t]
   \centering
   \footnotesize
   \scalebox{1.0}{
   \setlength{\tabcolsep}{2.5mm}{\begin{tabular}{c|c|c|c|c}
      \toprule[1pt]
      fusion module & mIoU & oIoU & mRR & rIoU \\
      \hline
      VLTF & 68.78 & 66.30 & 73.73 & 46.08 \\
      PWAN + LP & 69.72 & 70.40 & 66.58 & 44.58 \\
      \bottomrule[1pt]
   \end{tabular}}}
   \vspace{-2mm}
   \caption{\small Comparison between our proposed multi-modal fusion module VLTF and PWAN + LP in LAVT \cite{LAVT}. Both modules are applied in RefSegformer.}
   \label{tab:VLTF_LAVT}
\end{table}

\begin{figure}[!t]
	\centering
	\includegraphics[width=1.0\linewidth]{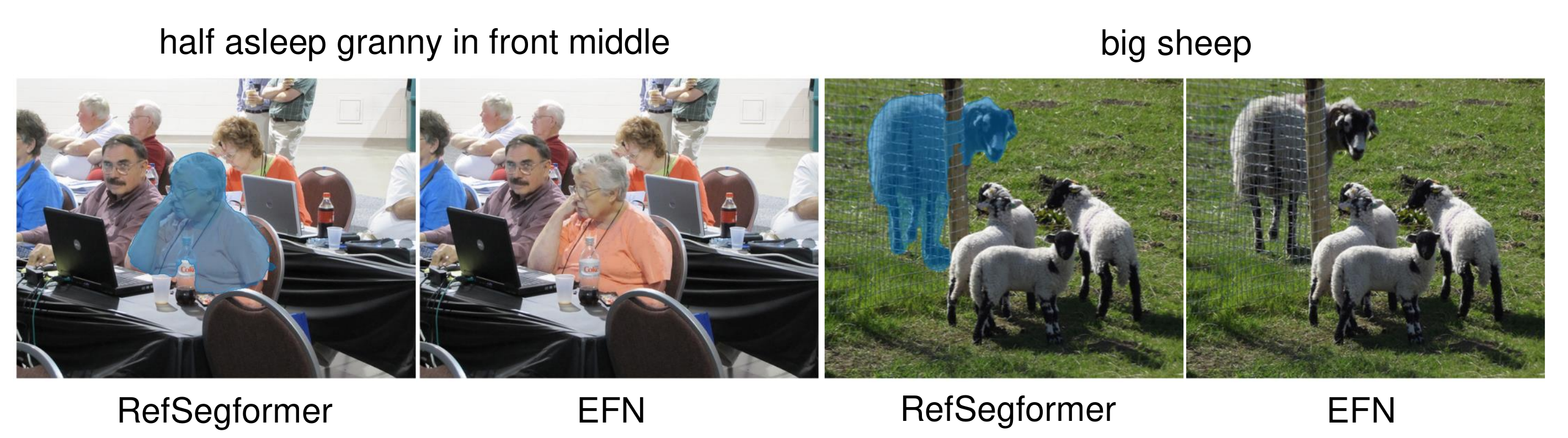}
	\caption{Visualization comparison between RefSegformer and EFN \cite{EFN}. All the expressions are \textbf{positive}.}
	\label{fig:sup_visual_EFN}
\end{figure}

\begin{figure*}[!t]
	\centering
	\includegraphics[width=1.0\linewidth]{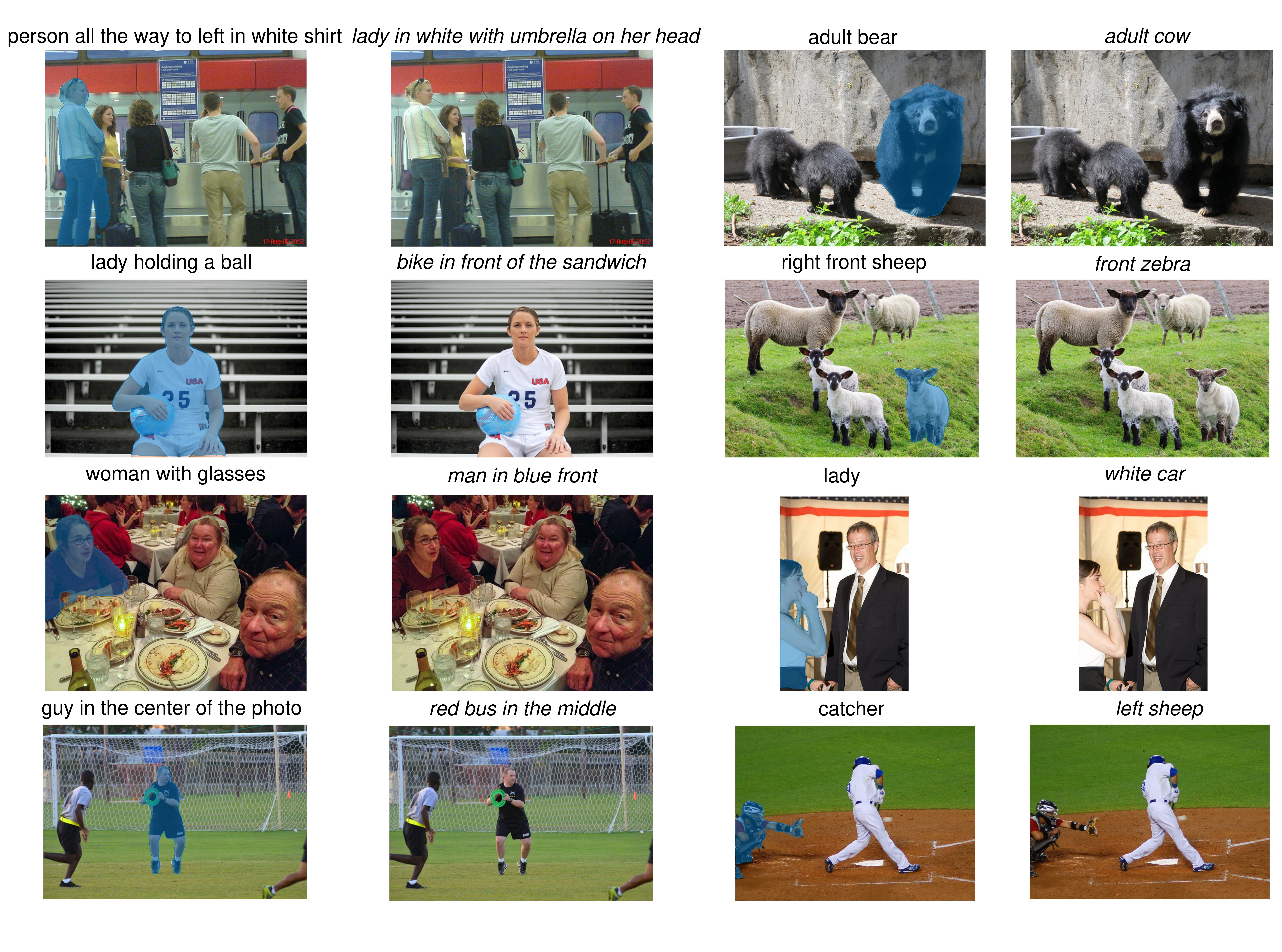}
	\caption{Visualization results of RefSegformer. Sentences in italics are negative sentences.}
	\label{fig:sup_visual}
\end{figure*}

\begin{figure*}[!t]
	\centering
	\includegraphics[width=1.0\linewidth]{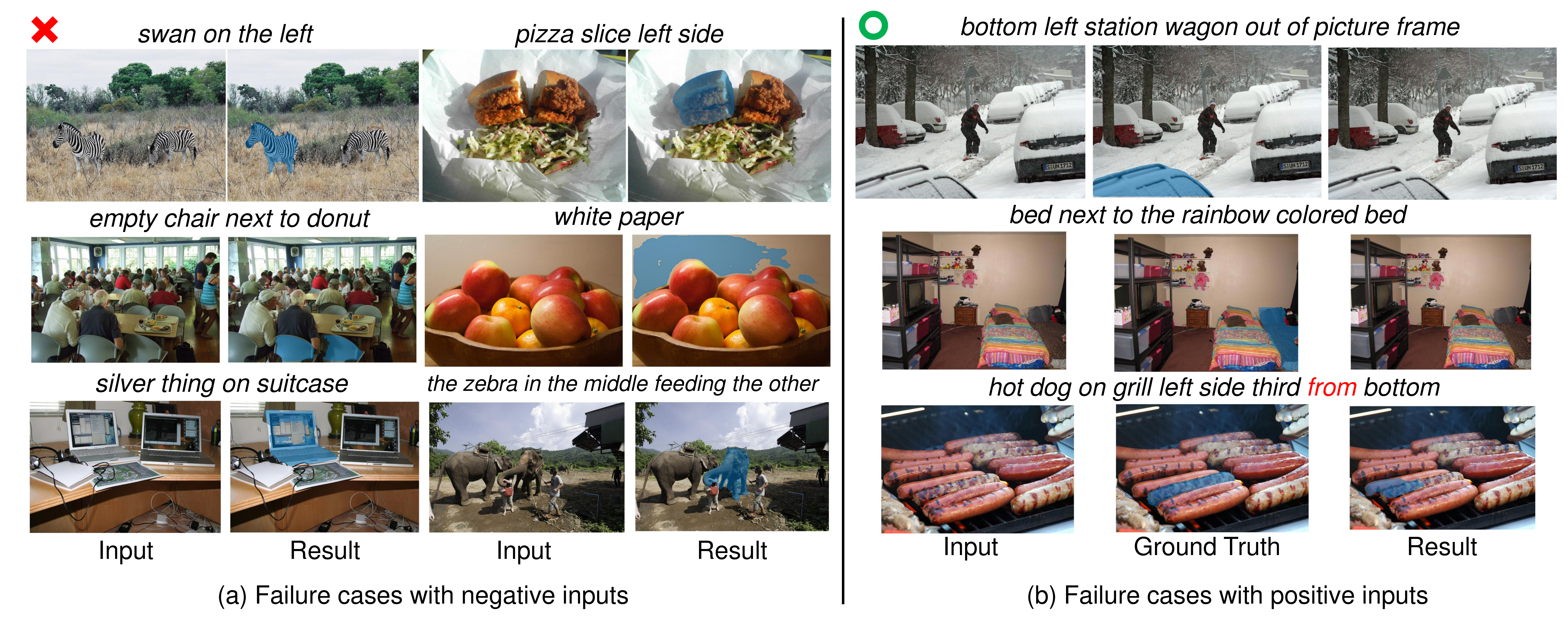}
	\caption{Visualization of failure cases.}
	\label{fig:failure_cases}
\end{figure*}

\noindent
\textbf{Comparison of fusion modules with LAVT.}
We compare VLTF with PWAN + LP (LAVT~\cite{LAVT}) in the architecture of RefSegformer. Results are shown in Tab.~\ref{tab:VLTF_LAVT}. RefSegformer adopting PWAN + LP achieves higher mIoU and oIoU metrics, while RefSegformer with VLTF module is higher in the mRR and rIoU metric. The reason is VLTF module contains blank tokens that do not interact with text features, thus disentangling the identification of positive and negative inputs. The results demonstrate VLTF is more suitable than PWAN + LP for the proposed R-RIS task.

\noindent
\textbf{Visualization Comparison with EFN.}
We present visualization results and compare the proposed RefSegformer with a previous work, EFN~\cite{EFN}. Fig.~\ref{fig:sup_visual_EFN} shows that the EFN model predicts no masks for positive sentence inputs, while RefSegformer predicts correct outputs. Both models are trained in the R-RIS task with negative inputs. EFN does not have modules designed specifically for the R-RIS task, thus the negative training samples harm the original architecture for the aligning of visual and linguistic features.

\noindent
\textbf{More Visual Results.}
We present more visualization results in Fig.~\ref{fig:sup_visual}. Our proposed RefSegformer can segment correct masks in RIS while identifying which sentence is wrong in R-RIS.

\noindent
\textbf{Failure Cases Analysis.}
We visualize some failure cases in Fig.~\ref{fig:failure_cases}. For negative inputs, when the object described by the input sentence is close to the objects in the image, the R-RIS model may incorrectly output masks. For example, in Fig.~\ref{fig:failure_cases}, ``zebra'' is identified as a ``swan'', and ``sandwich'' is considered as a piece of ``pizza''. For positive sentences, if the expressions are too complicated and contain infrequently used words like ``bottom left station wagon out of picture frame'', the model tends to output no masks, i.e., takes the input as negative. These results indicate that a better R-RIS model should be less sensitive to the localization words and more robust to the concepts with the similar shapes for the negative sentence. 

\subsection{Comparison With R2VOS}

In this section, we present the differences between our proposed R-RIS task and the recent R2VOS~\cite{R2VOS} work, which proposes a robust setting for video referring segmentation.

\noindent
\textbf{Different Settings.}
Unlike our proposed R-RIS, which is an extension on Referring \textbf{Image} Segmentation, R2VOS proposes a robust setting upon Referring \textbf{Video} Segmentation. The two tasks share certain similarities because they both focus on the problem that referring expressions may describe objects not existing in the image. However, the two tasks have no intersections since one is applied on the image level, while another is on the video level, like Referring Image Segmentation and Referring Video Segmentation. Moreover, our method is a generalized RIS task and mainly focus on the robustness of current RIS models.

\noindent
\textbf{Different Datasets.}
We build three validation datasets, including R-RefCOCO, R-RefCOCO+, and R-RefCOCOg, to evaluate an R-RIS model's performance. Each is built upon existing RIS datasets (RefCOCO, etc.) along with carefully generated negative sentences as negative inputs.
We believe a dataset that contains \textit{multiple negative sentences} for one reference can better test the robust ability of an R-RIS model. Thus, we generate 10 negative sentences for each reference, using a total of 5 generating methods (Details are in Sec. \ref{sec:more-exp}). In R2VOS, the proposed model is validated on the original RVOS datasets, with \textit{only one} negative sentence added per reference. The negative sentence is obtained by shuffling the original video set and constraining all negative text-video pairs unrelated, which is close to the ``Sentence'' method of our generating methods.

\noindent
\textbf{Different Metric.}
We propose two evaluation metrics for R-RIS, named rIoU and mRR. The rIoU metric measures the mask outputs of positive and negative sentences at the same time, and the mRR metric describes instance-level results of an R-RIS model. R2VOS introduces its metric for its task, named R. $R = 1 - \frac{\sum_{M \in V_{neg}}|M|}{\sum_{M \in V{pos}}|M|}$. According to the formula, 
$R$ becomes bigger when the model predicts smaller masks for the negative inputs compared to the positive inputs.
It measures the robustness but can not measure the segmentation quality for positive inputs.

{\small
\bibliographystyle{ieee_fullname}
\bibliography{egbib}

\begin{thebibliography}{10}\itemsep=-1pt

\bibitem{altindis2021benchmarking}
Said~Fahri Altindis, Yusuf Dalva, and Aysegul Dundar.
\newblock Benchmarking the robustness of instance segmentation models.
\newblock {\em arXiv preprint arXiv:2109.01123}, 2021.

\bibitem{botach2022end}
Adam Botach, Evgenii Zheltonozhskii, and Chaim Baskin.
\newblock End-to-end referring video object segmentation with multimodal
  transformers.
\newblock In {\em CVPR}, 2022.

\bibitem{detr}
Nicolas Carion, Francisco Massa, Gabriel Synnaeve, Nicolas Usunier, Alexander
  Kirillov, and Sergey Zagoruyko.
\newblock End-to-end object detection with transformers.
\newblock In {\em ECCV}. Springer, 2020.

\bibitem{chen2018domain}
Yuhua Chen, Wen Li, Christos Sakaridis, Dengxin Dai, and Luc Van~Gool.
\newblock Domain adaptive faster r-cnn for object detection in the wild.
\newblock In {\em CVPR}, 2018.

\bibitem{chen2019referring}
Yi-Wen Chen, Yi-Hsuan Tsai, Tiantian Wang, Yen-Yu Lin, and Ming-Hsuan Yang.
\newblock Referring expression object segmentation with caption-aware
  consistency.
\newblock {\em arXiv preprint arXiv:1910.04748}, 2019.

\bibitem{cheng2021maskformer}
Bowen Cheng, Alexander~G. Schwing, and Alexander Kirillov.
\newblock Per-pixel classification is not all you need for semantic
  segmentation.
\newblock In {\em NeurIPS}, 2021.

\bibitem{deng2009imagenet}
Jia Deng, Wei Dong, Richard Socher, Li-Jia Li, Kai Li, and Li Fei-Fei.
\newblock Imagenet: A large-scale hierarchical image database.
\newblock In {\em CVPR}, 2009.

\bibitem{BERT}
Jacob Devlin, Ming-Wei Chang, Kenton Lee, and Kristina Toutanova.
\newblock Bert: Pre-training of deep bidirectional transformers for language
  understanding.
\newblock {\em arXiv preprint arXiv:1810.04805}, 2018.

\bibitem{VLT}
Henghui Ding, Chang Liu, Suchen Wang, and Xudong Jiang.
\newblock Vision-language transformer and query generation for referring
  segmentation.
\newblock In {\em ICCV}, 2021.

\bibitem{ding2022vlt}
Henghui Ding, Chang Liu, Suchen Wang, and Xudong Jiang.
\newblock Vlt: Vision-language transformer and query generation for referring
  segmentation.
\newblock {\em ICCV}, 2022.

\bibitem{VIT}
Alexey Dosovitskiy, Lucas Beyer, Alexander Kolesnikov, Dirk Weissenborn,
  Xiaohua Zhai, Thomas Unterthiner, Mostafa Dehghani, Matthias Minderer, Georg
  Heigold, Sylvain Gelly, et~al.
\newblock An image is worth 16x16 words: Transformers for image recognition at
  scale.
\newblock {\em arXiv preprint arXiv:2010.11929}, 2020.

\bibitem{EFN}
Guang Feng, Zhiwei Hu, Lihe Zhang, and Huchuan Lu.
\newblock Encoder fusion network with co-attention embedding for referring
  image segmentation.
\newblock In {\em CVPR}, 2021.

\bibitem{gabeur2020mmt}
Valentin Gabeur, Chen Sun, Karteek Alahari, and Cordelia Schmid.
\newblock {Multi-modal Transformer for Video Retrieval}.
\newblock In {\em ECCV}, 2020.

\bibitem{geirhos2018generalisation}
Robert Geirhos, Carlos~RM Temme, Jonas Rauber, Heiko~H Sch{\"u}tt, Matthias
  Bethge, and Felix~A Wichmann.
\newblock Generalisation in humans and deep neural networks.
\newblock {\em NeurIPS}, 31, 2018.

\bibitem{Mask-RCNN}
Kaiming He, Georgia Gkioxari, Piotr Doll{\'a}r, and Ross Girshick.
\newblock Mask r-cnn.
\newblock In {\em ICCV}, 2017.

\bibitem{hendrycks2019benchmarking}
Dan Hendrycks and Thomas Dietterich.
\newblock Benchmarking neural network robustness to common corruptions and
  perturbations.
\newblock {\em arXiv preprint arXiv:1903.12261}, 2019.

\bibitem{hui2020linguistic}
Tianrui Hui, Si Liu, Shaofei Huang, Guanbin Li, Sansi Yu, Faxi Zhang, and
  Jizhong Han.
\newblock Linguistic structure guided context modeling for referring image
  segmentation.
\newblock In {\em ECCV}. Springer, 2020.

\bibitem{jain2021comprehensive}
Kanishk Jain and Vineet Gandhi.
\newblock Comprehensive multi-modal interactions for referring image
  segmentation.
\newblock In {\em Findings of the Association for Computational Linguistics:
  ACL 2022}, 2022.

\bibitem{LTS}
Ya Jing, Tao Kong, Wei Wang, Liang Wang, Lei Li, and Tieniu Tan.
\newblock Locate then segment: A strong pipeline for referring image
  segmentation.
\newblock In {\em CVPR}, 2021.

\bibitem{kamann2020benchmarking}
Christoph Kamann and Carsten Rother.
\newblock Benchmarking the robustness of semantic segmentation models.
\newblock In {\em CVPR}, 2020.

\bibitem{kamath2021mdetr}
Aishwarya Kamath, Mannat Singh, Yann LeCun, Ishan Misra, Gabriel Synnaeve, and
  Nicolas Carion.
\newblock Mdetr--modulated detection for end-to-end multi-modal understanding.
\newblock {\em ICCV}, 2021.

\bibitem{khoreva2018video}
Anna Khoreva, Anna Rohrbach, and Bernt Schiele.
\newblock Video object segmentation with language referring expressions.
\newblock In {\em ACCV}. Springer, 2018.

\bibitem{ReSTR}
Namyup Kim, Dongwon Kim, Cuiling Lan, Wenjun Zeng, and Suha Kwak.
\newblock Restr: Convolution-free referring image segmentation using
  transformers.
\newblock {\em arXiv preprint arXiv:2203.16768}, 2022.

\bibitem{li2023transformer}
Xiangtai Li, Henghui Ding, Wenwei Zhang, Haobo Yuan, Guangliang Cheng, Pang
  Jiangmiao, Kai Chen, Ziwei Liu, and Chen~Change Loy.
\newblock Transformer-based visual segmentation: A survey.
\newblock {\em arXiv pre-print}, 2023.

\bibitem{R2VOS}
Xiang Li, Jinglu Wang, Xiaohao Xu, Xiao Li, Yan Lu, and Bhiksha Raj.
\newblock R\^{} 2vos: Robust referring video object segmentation via relational
  multimodal cycle consistency.
\newblock {\em arXiv preprint arXiv:2207.01203}, 2022.

\bibitem{li2022panopticpartformer}
Xiangtai Li, Shilin Xu, Yibo Yang, Guangliang Cheng, Yunhai Tong, and Dacheng
  Tao.
\newblock Panoptic-partformer: Learning a unified model for panoptic part
  segmentation.
\newblock In {\em ECCV}, 2022.

\bibitem{li2023panopticpartformer++}
Xiangtai Li, Shilin Xu, Yibo Yang, Haobo Yuan, Guangliang Cheng, Yunhai Tong,
  Zhouchen Lin, Ming-Hsuan Yang, and Dacheng Tao.
\newblock Panopticpartformer++: A unified and decoupled view for panoptic part
  segmentation.
\newblock {\em arXiv preprint arXiv:2301.00954}, 2023.

\bibitem{li2020semantic}
Xiangtai Li, Ansheng You, Zhen Zhu, Houlong Zhao, Maoke Yang, Kuiyuan Yang,
  Shaohua Tan, and Yunhai Tong.
\newblock Semantic flow for fast and accurate scene parsing.
\newblock In {\em ECCV}, 2020.

\bibitem{Li2022SFNetFA}
Xiangtai Li, Jiangning Zhang, Yibo Yang, Guangliang Cheng, Kuiyuan Yang, Yu
  Tong, and Dacheng Tao.
\newblock Sfnet: Faster, accurate, and domain agnostic semantic segmentation
  via semantic flow.
\newblock {\em ArXiv}, abs/2207.04415, 2022.

\bibitem{li2022videoknet}
Xiangtai Li, Wenwei Zhang, Jiangmiao Pang, Kai Chen, Guangliang Cheng, Yunhai
  Tong, and Chen~Change Loy.
\newblock Video k-net: A simple, strong, and unified baseline for video
  segmentation.
\newblock In {\em CVPR}, 2022.

\bibitem{li2020gated}
Xiangtai Li, Houlong Zhao, Lei Han, Yunhai Tong, Shaohua Tan, and Kuiyuan Yang.
\newblock Gated fully fusion for semantic segmentation.
\newblock In {\em AAAI}, 2020.

\bibitem{li2021mail}
Zizhang Li, Mengmeng Wang, Jianbiao Mei, and Yong Liu.
\newblock Mail: A unified mask-image-language trimodal network for referring
  image segmentation.
\newblock {\em arXiv preprint arXiv:2111.10747}, 2021.

\bibitem{liang2022local}
Chen Liang, Wenguan Wang, Tianfei Zhou, Jiaxu Miao, Yawei Luo, and Yi Yang.
\newblock Local-global context aware transformer for language-guided video
  segmentation.
\newblock {\em arXiv preprint arXiv:2203.09773}, 2022.

\bibitem{liang2021clawcranenet}
Chen Liang, Yu Wu, Yawei Luo, and Yi Yang.
\newblock Clawcranenet: Leveraging object-level relation for text-based video
  segmentation.
\newblock {\em arXiv preprint arXiv:2103.10702}, 2021.

\bibitem{COCO_dataset}
Tsung-Yi Lin, Michael Maire, Serge Belongie, James Hays, Pietro Perona, Deva
  Ramanan, Piotr Doll{\'a}r, and C~Lawrence Zitnick.
\newblock Microsoft coco: Common objects in context.
\newblock In {\em ECCV}, 2014.

\bibitem{liu2022instance}
Chang Liu, Xudong Jiang, and Henghui Ding.
\newblock Instance-specific feature propagation for referring segmentation.
\newblock {\em TMM}, 2022.

\bibitem{liu2021pre}
Pengfei Liu, Weizhe Yuan, Jinlan Fu, Zhengbao Jiang, Hiroaki Hayashi, and
  Graham Neubig.
\newblock Pre-train, prompt, and predict: A systematic survey of prompting
  methods in natural language processing.
\newblock {\em arXiv preprint arXiv:2107.13586}, 2021.

\bibitem{CMPC}
Si Liu, Tianrui Hui, Shaofei Huang, Yunchao Wei, Bo Li, and Guanbin Li.
\newblock Cross-modal progressive comprehension for referring segmentation.
\newblock {\em PAMI}, 2021.

\bibitem{swin}
Ze Liu, Yutong Lin, Yue Cao, Han Hu, Yixuan Wei, Zheng Zhang, Stephen Lin, and
  Baining Guo.
\newblock Swin transformer: Hierarchical vision transformer using shifted
  windows.
\newblock In {\em ICCV}, 2021.

\bibitem{ADAMW}
Ilya Loshchilov and Frank Hutter.
\newblock Decoupled weight decay regularization, 2017.

\bibitem{CGAN}
Gen Luo, Yiyi Zhou, Rongrong Ji, Xiaoshuai Sun, Jinsong Su, Chia-Wen Lin, and
  Qi Tian.
\newblock Cascade grouped attention network for referring expression
  segmentation.
\newblock In {\em ACM}, 2020.

\bibitem{MCN}
Gen Luo, Yiyi Zhou, Xiaoshuai Sun, Liujuan Cao, Chenglin Wu, Cheng Deng, and
  Rongrong Ji.
\newblock Multi-task collaborative network for joint referring expression
  comprehension and segmentation.
\newblock In {\em CVPR}, 2020.

\bibitem{RefCOCOg}
Junhua Mao, Jonathan Huang, Alexander Toshev, Oana Camburu, Alan~L Yuille, and
  Kevin Murphy.
\newblock Generation and comprehension of unambiguous object descriptions.
\newblock In {\em CVPR}, 2016.

\bibitem{RefCOCOg2}
Varun~K Nagaraja, Vlad~I Morariu, and Larry~S Davis.
\newblock Modeling context between objects for referring expression
  understanding.
\newblock In {\em ECCV}. Springer, 2016.

\bibitem{neyshabur2017exploring}
Behnam Neyshabur, Srinadh Bhojanapalli, David McAllester, and Nati Srebro.
\newblock Exploring generalization in deep learning.
\newblock {\em NeurIPS}, 30, 2017.

\bibitem{CLIP}
Alec Radford, Jong~Wook Kim, Chris Hallacy, Aditya Ramesh, Gabriel Goh,
  Sandhini Agarwal, Girish Sastry, Amanda Askell, Pamela Mishkin, Jack Clark,
  et~al.
\newblock Learning transferable visual models from natural language
  supervision.
\newblock In {\em ICML}. PMLR, 2021.

\bibitem{rice2020overfitting}
Leslie Rice, Eric Wong, and Zico Kolter.
\newblock Overfitting in adversarially robust deep learning.
\newblock In {\em ICML}, 2020.

\bibitem{sakaridis2019guided}
Christos Sakaridis, Dengxin Dai, and Luc~Van Gool.
\newblock Guided curriculum model adaptation and uncertainty-aware evaluation
  for semantic nighttime image segmentation.
\newblock In {\em ICCV}, 2019.

\bibitem{sakaridis2018semantic}
Christos Sakaridis, Dengxin Dai, and Luc Van~Gool.
\newblock Semantic foggy scene understanding with synthetic data.
\newblock {\em IJCV}, 2018.

\bibitem{seo2020urvos}
Seonguk Seo, Joon-Young Lee, and Bohyung Han.
\newblock Urvos: Unified referring video object segmentation network with a
  large-scale benchmark.
\newblock In {\em ECCV}. Springer, 2020.

\bibitem{deit_vit}
Hugo Touvron, Matthieu Cord, Matthijs Douze, Francisco Massa, Alexandre
  Sablayrolles, and Herv{\'e} J{\'e}gou.
\newblock Training data-efficient image transformers \& distillation through
  attention.
\newblock In {\em ICML}. PMLR, 2021.

\bibitem{vasiljevic2016examining}
Igor Vasiljevic, Ayan Chakrabarti, and Gregory Shakhnarovich.
\newblock Examining the impact of blur on recognition by convolutional
  networks.
\newblock {\em arXiv preprint arXiv:1611.05760}, 2016.

\bibitem{CRIS}
Zhaoqing Wang, Yu Lu, Qiang Li, Xunqiang Tao, Yandong Guo, Mingming Gong, and
  Tongliang Liu.
\newblock Cris: Clip-driven referring image segmentation.
\newblock {\em arXiv preprint arXiv:2111.15174}, 2021.

\bibitem{wu2022language}
Jiannan Wu, Yi Jiang, Peize Sun, Zehuan Yuan, and Ping Luo.
\newblock Language as queries for referring video object segmentation.
\newblock In {\em CVPR}, 2022.

\bibitem{wu2023betrayed}
Jianzong Wu, Xiangtai Li, Henghui Ding, Xia Li, Guangliang Cheng, Yunhai Tong,
  and Chen~Change Loy.
\newblock Betrayed by captions: Joint caption grounding and generation for open
  vocabulary instance segmentation.
\newblock {\em ICCV}, 2023.

\bibitem{xie2021segformer}
Enze Xie, Wenhai Wang, Zhiding Yu, Anima Anandkumar, Jose~M Alvarez, and Ping
  Luo.
\newblock Segformer: Simple and efficient design for semantic segmentation with
  transformers.
\newblock In {\em NeurIPS}, 2021.

\bibitem{xu2022fashionformer}
Shilin Xu, Xiangtai Li, Jingbo Wang, Guangliang Cheng, Yunhai Tong, and Dacheng
  Tao.
\newblock Fashionformer: A simple, effective and unified baseline for human
  fashion segmentation and recognition.
\newblock {\em ECCV}, 2022.

\bibitem{yamada2022does}
Yutaro Yamada and Mayu Otani.
\newblock Does robustness on imagenet transfer to downstream tasks?
\newblock In {\em CVPR}, 2022.

\bibitem{BUSNet}
Sibei Yang, Meng Xia, Guanbin Li, Hong-Yu Zhou, and Yizhou Yu.
\newblock Bottom-up shift and reasoning for referring image segmentation.
\newblock In {\em CVPR}, 2021.

\bibitem{LAVT}
Zhao Yang, Jiaqi Wang, Yansong Tang, Kai Chen, Hengshuang Zhao, and Philip~HS
  Torr.
\newblock Lavt: Language-aware vision transformer for referring image
  segmentation.
\newblock In {\em CVPR}, 2022.

\bibitem{ye2019cross}
Linwei Ye, Mrigank Rochan, Zhi Liu, and Yang Wang.
\newblock Cross-modal self-attention network for referring image segmentation.
\newblock In {\em CVPR}, 2019.

\bibitem{yu2018mattnet}
Licheng Yu, Zhe Lin, Xiaohui Shen, Jimei Yang, Xin Lu, Mohit Bansal, and
  Tamara~L Berg.
\newblock Mattnet: Modular attention network for referring expression
  comprehension.
\newblock In {\em CVPR}, 2018.

\bibitem{RefCOCO}
Licheng Yu, Patrick Poirson, Shan Yang, Alexander~C Berg, and Tamara~L Berg.
\newblock Modeling context in referring expressions.
\newblock In {\em ECCV}. Springer, 2016.

\bibitem{zhang2023rethinking}
Jiangning Zhang, Xiangtai Li, Jian Li, Liang Liu, Zhucun Xue, Boshen Zhang,
  Zhengkai Jiang, Tianxin Huang, Yabiao Wang, and Chengjie Wang.
\newblock Rethinking mobile block for efficient neural models.
\newblock {\em ICCV}, 2023.

\bibitem{zhang2022eatformer}
Jiangning Zhang, Xiangtai Li, Yabiao Wang, Chengjie Wang, Yibo Yang, Yong Liu,
  and Dacheng Tao.
\newblock Eatformer: Improving vision transformer inspired by evolutionary
  algorithm.
\newblock {\em arXiv preprint arXiv:2206.09325}, 2022.

\bibitem{zhang2021analogous}
Jiangning Zhang, Chao Xu, Jian Li, Wenzhou Chen, Yabiao Wang, Ying Tai, Shuo
  Chen, Chengjie Wang, Feiyue Huang, and Yong Liu.
\newblock Analogous to evolutionary algorithm: Designing a unified sequence
  model.
\newblock {\em NeurIPS}, 2021.

\bibitem{zhou2022understanding}
Daquan Zhou, Zhiding Yu, Enze Xie, Chaowei Xiao, Animashree Anandkumar, Jiashi
  Feng, and Jose~M Alvarez.
\newblock Understanding the robustness in vision transformers.
\newblock In {\em ICML}, 2022.

\bibitem{deformable-DETR}
Xizhou Zhu, Weijie Su, Lewei Lu, Bin Li, Xiaogang Wang, and Jifeng Dai.
\newblock Deformable detr: Deformable transformers for end-to-end object
  detection.
\newblock {\em arXiv preprint arXiv:2010.04159}, 2020.

\end{thebibliography}
}

\end{document}